\documentclass{article} 
\usepackage[final]{colm2026_conference}

\usepackage{microtype}
\usepackage{hyperref}
\usepackage{url}
\usepackage{booktabs}
\usepackage{xltabular} 
\usepackage{placeins}  
\usepackage{multirow}
\usepackage{tabularx}
\usepackage{array}
\usepackage{graphicx}
\newcolumntype{Y}{>{\raggedright\arraybackslash}X}

\usepackage{lineno}
\usepackage{amsmath}
\usepackage{amssymb}
\usepackage{enumitem}
\usepackage{etoolbox}
\usepackage{xcolor}
\usepackage{colortbl}
\usepackage{tikz}
\usetikzlibrary{arrows.meta, positioning, fit, backgrounds, calc, shapes.geometric}

\definecolor{mBlue}{HTML}{4C6A92}
\definecolor{mRust}{HTML}{B5715B}
\definecolor{mSage}{HTML}{7A9A7E}
\definecolor{mGold}{HTML}{C9A66B}
\definecolor{mPlum}{HTML}{8A7096}
\definecolor{mSlate}{HTML}{5A6674}
\definecolor{mInk}{HTML}{2E3440}
\definecolor{hdrBG}{HTML}{EAEEF4}
\definecolor{rowBG}{HTML}{F7F9FB}
\definecolor{keyBG}{HTML}{E3EBF5}
\definecolor{goodBG}{HTML}{E8F0EA}
\definecolor{warnBG}{HTML}{F8EDE8}
\definecolor{noteBG}{HTML}{F4F1E9}

\newcommand{\hdr}{\rowcolor{hdrBG}}
\newcommand{\rw}{\rowcolor{rowBG}}
\newcommand{\keycell}[1]{\cellcolor{keyBG}#1}
\newcommand{\goodcell}[1]{\cellcolor{goodBG}#1}
\newcommand{\warncell}[1]{\cellcolor{warnBG}#1}

\newcommand{\TableFontSize}{\scriptsize}

\AtBeginEnvironment{equation}{\footnotesize}
\AtBeginEnvironment{equation*}{\footnotesize}
\AtBeginEnvironment{align}{\footnotesize}
\AtBeginEnvironment{align*}{\footnotesize}
\AtBeginEnvironment{gather}{\footnotesize}
\AtBeginEnvironment{gather*}{\footnotesize}
\AtBeginEnvironment{multline}{\footnotesize}
\AtBeginEnvironment{multline*}{\footnotesize}
\AtBeginEnvironment{flalign}{\footnotesize}
\AtBeginEnvironment{flalign*}{\footnotesize}

\AtBeginEnvironment{table}{\TableFontSize}
\AtBeginEnvironment{table*}{\TableFontSize}

\definecolor{darkblue}{rgb}{0, 0, 0.5}
\hypersetup{colorlinks=true, citecolor=darkblue, linkcolor=darkblue, urlcolor=darkblue}

\usepackage{amsthm}

\usepackage{csquotes}
\MakeOuterQuote{"}

\makeatletter
\patchcmd{\@maketitle}{{\Large\bf \@title\par}}{{\Large\bf\centering \@title\par}}{}{%
  \PackageWarning{colm-local}{could not centre title}}
\makeatother

\title{\textit{Actions Speak Louder than Words}: Measuring Cross-Lingual Policy Retention in Tool-Using Agents}

\author{%
\begin{minipage}{\dimexpr\textwidth-2\tabcolsep\relax}
\centering
\normalsize
\textbf{Sourabrata Mukherjee} \qquad \textbf{Kalika Bali} \qquad \textbf{Sunayana Sitaram}\\[3pt]
\normalfont Microsoft Research India\\[1.5pt]
\normalfont\texttt{t-somukherje@microsoft.com}
\end{minipage}%
}

\begin{document}

\ifcolmsubmission
\linenumbers
\fi

\maketitle

\begin{abstract}
When a tool-using agent is given the same task in a different language, does it still take the same
steps? Multilingual evaluation rarely asks, because it compares final answers and throws the
intermediate actions away. For an agent those actions are the product: the route fixes cost and
latency, decides how the system fails, and is the only part of its behaviour a reviewer can audit,
so two language versions that agree on every answer can still differ in price, in failure mode, and
in whether a safeguard written against one of them holds for the other. We therefore make the
action policy itself the measured object, across 8 models, 6 parallel benchmarks and 41 languages
(2.38M agent rollouts). The naive measurement does not work, because five confounds sit between raw
trace similarity and any defensible claim, each large enough on its own to change a conclusion:
short traces score higher than long ones, empty traces score perfectly, unrelated traces already
agree by chance more than half the time, the gap is capped by each model's own reproducibility, and,
worst of all, a model asked the same question twice in one language does not answer it the same way,
so there is no reference point to measure against. We rebuild the measurement to remove all five,
and every correction makes the effect larger rather than smaller.
Cross-lingual divergence then proves structural rather than sampling noise: it survives greedy
decoding in every cell we test and stays flat as temperature rises, even as models become far less
self-consistent. Normalising by each model's own reproducibility, four very different frontier
models converge on nearly the same value \emph{under greedy decoding}, each keeping 71--73\% of its
action policy when the language changes, and which model it is explains only 5.7\% of the
variance. Below roughly 10B
parameters that regularity breaks down, and the apparent ordering among smaller models is largely an
artifact of a chance floor that we measure by permutation rather than assume. Asking why, we find
that agents route non-English tasks through English, that this pivot is causally load-bearing, with
a prediction registered in advance and confirmed across four models, and that models will not
abandon it when instructed to. Finally, a single trace-extraction regex, not the model, manufactured
an apparent multilingual failure: two worked examples raise one model's measured accuracy
twenty-sixfold while its accuracy on readable outputs barely moves. Code, prompts, both benchmark
suites and all 2.38M per-rollout traces are released.\footnote{All artifacts are available from the
\href{https://aka.ms/multilingual_agents_evaluation}{project repository}. Key terms:
Table~\ref{tab:keyterms}, Appendix~\ref{app:keyterms}.}
\end{abstract}

\section{Introduction}
\label{sec:intro}

Language models are increasingly deployed as \emph{agents}: they break a task into steps, call
tools, and assemble a result~\citep{yao2023react, schick2023toolformer, parisi2022talm,
qin2024toolllm}. Ship one multilingually and the question that matters is not only whether it
answers correctly in Hindi, but whether it \emph{does the same thing} in Hindi as in English:
whether it retrieves before it computes, whether it translates first, whether it takes three steps
or eight.

That difference is not cosmetic, because for an agent the route is the product. A model that
translates a passage before answering spends tokens and latency that the English arm never spends,
so the identical request costs more in one language than in another. A different route also fails
differently: an error introduced by that translation step exists only on the non-English path and
will never surface in an English regression test. And the route is the part of an agent's behaviour
that can actually be governed, since tool permissions, rate limits, escalation rules and audit
trails are all written against the sequence of actions a system is expected to take, so a policy
validated on the English trace does not describe what the system does in Hindi. None of this means
a different language should always produce an identical route; sometimes it should not. It means
the difference has to be measurable, and today it is not measured at all, because answer-level
parity is compatible with any amount of it.

\begin{figure}[!tb]
\centering
\begin{tikzpicture}[
  x=1mm, y=1mm,
  every node/.style={align=center, inner sep=1.4pt},
  panel/.style   ={rounded corners=2.4pt, draw=mSlate!22, fill=rowBG, line width=0.4pt},
  ptitle/.style  ={font=\fontsize{7}{8}\selectfont\bfseries, text=mSlate, anchor=west},
  task/.style    ={rounded corners=2pt, draw=mSlate!45, fill=white, line width=0.4pt,
                   font=\fontsize{6.4}{7.4}\selectfont, inner sep=2.4pt},
  arm/.style     ={font=\fontsize{6.6}{7.4}\selectfont, text=mInk},
  rep/.style     ={font=\fontsize{5.8}{6.6}\selectfont, text=mSlate!90, anchor=east},
  tok/.style     ={rounded corners=1.2pt, text=white, minimum height=3.3mm,
                   minimum width=8.6mm, font=\fontsize{5.3}{6}\selectfont, inner sep=0.6pt},
  chip/.style    ={rounded corners=1.6pt, draw=mRust!45, fill=white, line width=0.4pt},
  chipc/.style   ={chip, draw=mRust!85, line width=0.55pt, dash pattern=on 1.3pt off 1.1pt},
  ctag/.style    ={anchor=west, font=\fontsize{6}{7}\selectfont},
  cval/.style    ={anchor=east, font=\fontsize{6}{7}\selectfont, text=mSlate},
  note/.style    ={anchor=west, font=\fontsize{5.9}{6.8}\selectfont, text=mSlate},
  ar/.style      ={-{Stealth[length=2.6pt,width=2.2pt]}, draw=mSlate!70, line width=0.45pt},
  win/.style     ={draw=mSage!95, line width=0.8pt, line cap=round, line join=round},
  crs/.style     ={{Stealth[length=2.6pt,width=2.2pt]}-{Stealth[length=2.6pt,width=2.2pt]},
                   draw=mRust!95, line width=0.8pt, dash pattern=on 1.5pt off 1.2pt},
]

\node[panel, fit={(1,-2.5) (80,-43.5)}, inner sep=0pt] {};
\node[ptitle] at (3.5,-5.2) {What we measure};

\node[task] at (40,-10.6) {one canonical task, rendered in 41 languages};
\draw[ar] (33,-12.7) -- (27,-15.4);
\draw[ar] (47,-12.7) -- (59,-15.4);
\node[arm] at (26.7,-17.3) {English};
\node[arm] at (58.7,-17.3) {Hindi};

\node[rep] at (8.4,-21.6) {run 1};
\node[tok, fill=mSlate] at (17.5,-21.6) {Search};
\node[tok, fill=mGold]  at (26.7,-21.6) {Calc};
\node[tok, fill=mBlue]  at (35.9,-21.6) {Finish};
\node[rep] at (8.4,-27.0) {run 2};
\node[tok, fill=mSlate] at (17.5,-27.0) {Search};
\node[tok, fill=mGold]  at (26.7,-27.0) {Calc};
\node[tok, fill=mBlue]  at (35.9,-27.0) {Finish};

\node[tok, fill=mRust]  at (49.5,-21.6) {Translate};
\node[tok, fill=mSlate] at (58.7,-21.6) {Search};
\node[tok, fill=mBlue]  at (67.9,-21.6) {Finish};
\node[tok, fill=mRust]  at (49.5,-27.0) {Translate};
\node[tok, fill=mGold]  at (58.7,-27.0) {Calc};
\node[tok, fill=mBlue]  at (67.9,-27.0) {Finish};

\draw[win] (72.6,-21.6) -- (75.6,-21.6) -- (75.6,-27.0) -- (72.6,-27.0);
\draw[win] (12.8,-21.6) -- (10.4,-21.6) -- (10.4,-27.0) -- (12.8,-27.0);
\draw[crs] (40.9,-22.3) -- (44.5,-26.3);

\node[anchor=west, font=\fontsize{5.9}{6.8}\selectfont] at (3.5,-31.9)
  {\textcolor{mSage}{\rule[0.55ex]{3.4mm}{0.7pt}}\;$I_{\text{within}}$, same language
   \quad\textcolor{mRust}{\rule[0.55ex]{1.2mm}{0.7pt}\,\rule[0.55ex]{1.2mm}{0.7pt}}\;$I_{\text{cross}}$, across languages};
\node[note] at (3.5,-34.9) {both compare run 1 with run 2, so only the language differs};

\node[anchor=west, font=\fontsize{7.6}{8.6}\selectfont] at (3.5,-38.6)
  {$\tilde{I} \;=\; I_{\text{cross}}\,/\,I_{\text{within}}$};
\node[note] at (3.5,-41.6) {the share of self-consistency that survives a change of language};

\node[panel, fit={(84,-2.5) (139,-43.5)}, inner sep=0pt] {};
\node[ptitle] at (86.5,-5.2) {Why the naive version fails};

\foreach \y/\st in {-10.4/chip, -15.2/chipc, -20.0/chip, -24.8/chip, -29.6/chipc}
  {\draw[\st] (86.5,\y-2.15) rectangle (136.5,\y+2.15);}
\node[ctag] at (88.2,-10.4) {\textbf{C1} no baseline};      \node[cval] at (134.8,-10.4) {a model is not itself};
\node[ctag] at (88.2,-15.2) {\textbf{C2} trace length};     \node[cval] at (134.8,-15.2) {causal, $6$--$7$ pts};
\node[ctag] at (88.2,-20.0) {\textbf{C3} empty traces};     \node[cval] at (134.8,-20.0) {score $1.0$};
\node[ctag] at (88.2,-24.8) {\textbf{C4} ceiling};          \node[cval] at (134.8,-24.8) {$r = +0.97$};
\node[ctag] at (88.2,-29.6) {\textbf{C5} chance floor};     \node[cval] at (134.8,-29.6) {measured, $c \approx 0.56$};

\node[note] at (86.5,-33.5) {dotted: established causally, not assumed};

\node[anchor=west, font=\fontsize{7.2}{8.2}\selectfont] at (86.5,-38.0)
  {$+0.063 \;\rightarrow\; +0.086 \;\rightarrow\; \mathbf{+0.207}$};
\node[note, text=mSage!72!black] at (86.5,-41.4)
  {every correction makes the effect \textbf{larger}};

\node[font=\fontsize{6.2}{7.2}\selectfont, text=mSlate] at (70,-46.6)
  {\textbf{8} models \;$\cdot$\; \textbf{6} parallel benchmarks \;$\cdot$\; \textbf{41} languages
   \;$\cdot$\; \textbf{505} cells \;$\cdot$\; \textbf{2.38M} rollouts};

\end{tikzpicture}
\caption{Semantically identical tasks in 41 languages pass through one fixed symbolic tool-use
scaffold, and every cell is generated twice so that same-language and cross-language agreement are
estimated the same way. The traces are illustrative: the Hindi arm opens with the English pivot,
and the two runs of a single language do not match each other either, which is why a same-language
baseline is needed before any cross-language number means anything. Five confounds sit between raw
trace similarity and a defensible claim, four of them large enough on their own to reverse a sign
or a ranking.}
\label{fig:overview}
\end{figure}
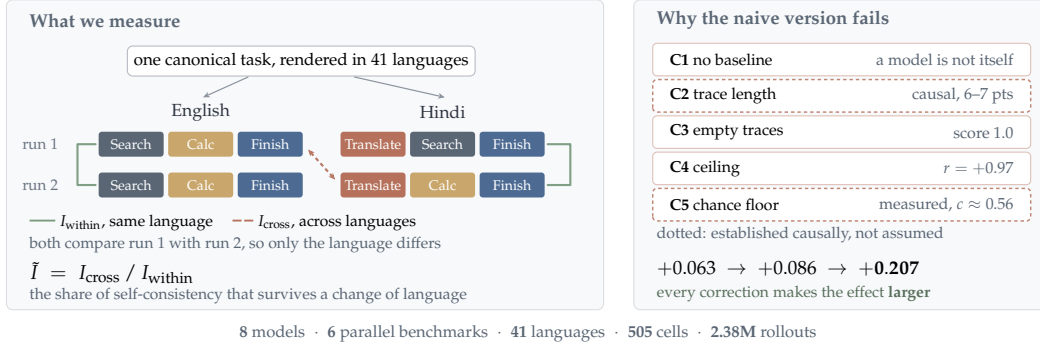

Multilingual evaluation has almost no purchase here. Standard suites compare final
outputs~\citep{hu2020xtreme, ahuja2023mega, liang2023helm}, and agentic benchmarks are
overwhelmingly English~\citep{liu2024agentbench, mialon2024gaia}, so the intermediate policy is
discarded on both sides. We make that policy the measured object and ask: \textbf{do multilingual
tool-using models execute the same action policy for semantically identical tasks across
languages?} Figure~\ref{fig:overview} summarises how we answer it.

What is hard here is not collecting the traces but interpreting them: rendering a task in many
languages, extracting the traces and averaging their similarity produces a number that cannot be
trusted, for five separate reasons.
Four are metric artifacts: short traces score higher than long ones, empty traces score perfectly,
unrelated traces already agree more than half the time by chance, and the gap is bounded by a
model's own reproducibility. The fifth is deeper. There is no baseline, because a model asked the
same question twice in its own language does not answer the same way, and without that reference
point the quantity is not identifiable at all.

Our protocol fixes this. Under a fixed scaffold with a shared five-tool alphabet, a model emits an
executed action trace alongside its answer, and the trace is what we compare. We generate every cell
\emph{twice} under identical decoding, task set and token budget, varying only the serving seed.
Same-language agreement $I_{\text{within}}$ pairs the two replicates in one language; cross-language
agreement $I_{\text{cross}}$ pairs them across two. Both sides are cross-seed, so decoding noise
enters identically and language is the only difference. \textbf{Normalised policy retention} is
$\tilde{I} = I_{\text{cross}}/I_{\text{within}}$, the share of a model's own reproducibility that
survives a change of language. Every comparison is length-matched in both directions and accepted
only when both agree in sign; pairs are dropped whenever either trace is empty; intervals are
task-level bootstraps. The estimator is set out in full in Appendix~\ref{app:protocol}, the five
confounds with their measured cost in Appendix~\ref{app:confounds}, and the design checks in
Appendix~\ref{app:design-checks}.

\begin{figure}[!tb]
\centering
\includegraphics[width=\textwidth]{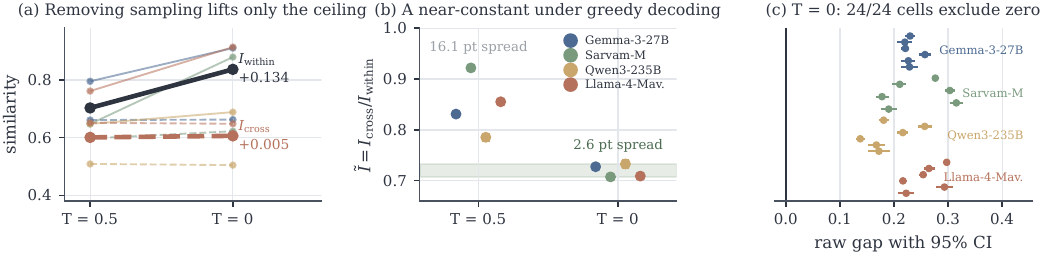}
\caption{The central result. (a) Greedy decoding raises $I_{\text{within}}$ by $0.134$ but moves
$I_{\text{cross}}$ by $0.005$; thin lines are models, thick lines the pooled mean. (b) Dividing by
each model's ceiling collapses a 13.6-point spread at $T{=}0.5$ into a 2.6-point band at $T{=}0$,
with task-level bootstrap 95\% intervals. (c) All 24 cells at $T{=}0$ are positive, with every
interval on the raw gap excluding zero.}
\label{fig:headline}
\end{figure}

Corrected this way, the effect grows rather than shrinks. Every correction we apply makes it
larger, never smaller. Under greedy decoding the gap is positive in all 24 cells, with every
bootstrap interval on the raw gap excluding zero (Figure~\ref{fig:headline}), and a five-point temperature ladder shows why
sampling had been hiding it: cross-lingual agreement barely moves as temperature rises, while
self-consistency falls away sharply.

Dividing by that ceiling gives the paper's central result. Under greedy decoding, four models that
share almost nothing, an order of magnitude of scale, two architectures and entirely different
training mixtures, converge on nearly the same value: each retains \textbf{71--73\%} of its own
action policy when the language changes. The regularity holds per cell rather than only per model, and across all 24 of
them model identity explains barely any of the variance (5.7\%). Extending the study to three
smaller and other-vendor models shows where this regularity ends, and why the ordering among small
models cannot be read off the raw metric at all (\S\ref{sec:estimand}).

Asking why it happens points to English. Agents route non-English tasks through an English pivot:
\texttt{Translate} is the most-used tool in every adapted benchmark, and reasoning text is
$\approx$99\% ASCII even on Devanagari input. Removing the translation tool lowers length-matched
agreement in proportion to how much a model uses it, and mandating it helps in a
\textbf{pre-registered} ordering across four models, predicted in advance from each model's measured
head-room. The pivot survives a direct instruction to abandon it, at under 1\% compliance. This
connects interpretability findings about latent English processing~\citep{wendler2024llamas,
zhao2024multilingualism} to a causal test at the level of executed actions. Separately, a single
trace-extraction regex, not the model, manufactured an apparent multilingual failure in
GPT-OSS-120B, which we take apart in \S\ref{sec:correctness}.

The work sits between two literatures that rarely meet. Agentic benchmarks built on ReAct-style
scaffolds~\citep{yao2023react, schick2023toolformer, liu2024agentbench, mialon2024gaia} score
\emph{outcomes} and are overwhelmingly English, while cross-lingual suites~\citep{hu2020xtreme,
ahuja2023mega, liang2023helm} compare final predictions and multilingual chain-of-thought
work~\citep{shi2023multilingualcot, muennighoff2023crosslingual} compares free-form \emph{text},
which is hard to align. We make the executed trace the measured object, which a shared symbolic
action alphabet renders comparable. Closest in spirit, \citet{qi2023crosslingual} ask whether models
retrieve the same \emph{facts} across languages; we ask it of \emph{procedures}, and add the matched
baseline without which the quantity is not identifiable (Appendix~\ref{app:formal}, with an extended
comparison in Appendix~\ref{app:related}).

We contribute a ceiling-corrected estimand for cross-lingual policy retention, and a measurement
protocol that prices each of five confounds on its own data, two of them established causally rather
than assumed (Appendices~\ref{app:protocol} and~\ref{app:confounds}). We apply it at a scale that
lets the question be answered rather than gestured at: 2.38M rollouts over 505 cells, spanning 8
models, 41 languages, five temperatures and six interventions (\S\ref{sec:setup}). We show the
retention regularity is real, locate its boundary, and demonstrate that the apparent ordering among
models below that boundary is a metric artifact (\S\ref{sec:estimand}). And we establish the
mechanism causally, with the key prediction registered in advance (\S\ref{sec:mechanism}).

\section{Experimental Setup}
\label{sec:setup}

Comparing policies across languages requires $x^{(\ell_i)}(z)$ and $x^{(\ell_j)}(z)$ to be the
\emph{same} task, a stronger condition than it looks and one that eliminates most multilingual
corpora. En--X bitext pairs each language with English but not with the others, so row $i$ of the
Hindi arm and row $i$ of the Bengali arm are unrelated sentences. Of the corpora we screened, five
fail this test outright, and a sixth stores its per-language configurations in different row
orders, so that an index join matches only half the items. The suite we keep is six benchmarks on
verified alignment keys (FLORES-200~\citep{nllb2022}, XQuAD~\citep{xquad2020},
XNLI~\citep{xnli2018}, Belebele~\citep{belebele2024}, XCOPA~\citep{xcopa2020} and a purpose-built
synthetic benchmark): \textbf{5{,}776 tasks}, \textbf{41 languages} across 17 families and 16
scripts, and 809 (benchmark, language-pair) comparisons over 428 distinct pairs. We call the five repurposed public benchmarks \emph{adapted} and
the purpose-built suite \emph{synthetic}. Per-benchmark statistics, the language inventory and
every alignment key are in Appendix~\ref{app:data}.

Five instruction-tuned models form the main suite, spanning scale, architecture and language
specialisation: \textbf{Gemma-3-27B}~\citep{gemma3_2025} and \textbf{Sarvam-M} (24B,
Indic-specialised) are dense; \textbf{Qwen3-235B-A22B}~\citep{qwen3_2025} and
\textbf{Llama-4-Maverick} (17B active, 128 experts) are mixture-of-experts;
\textbf{GPT-OSS-120B}~\citep{gptoss2025} is the fifth. To locate the boundary of the regularity in
\S\ref{sec:estimand} we add \textbf{Gemma-3-4B} (same family and recipe, $6.75\times$ smaller,
isolating scale), \textbf{Qwen3-8B} (same family, different training mix) and
\textbf{Aya-Expanse-8B}~\citep{ayaexpanse2024} (an independent vendor, explicitly multilingual
post-training), giving
\textbf{eight models in total}. Every task--language instance uses an identical scaffold requiring
\texttt{Thought:}\,/\,\texttt{Action: ToolName(args)} steps ending in \texttt{Finish}, with traces
extracted by a single regex. Tools are \emph{symbolic}: calls are parsed and compared but never
executed, so we study induced tool-use policy rather than grounded execution. Unless stated
otherwise decoding is $T{=}0.5$, \texttt{max\_tokens}$=$4096 and an iteration cap of 10; serving is
vLLM~\citep{kwon2023vllm} on $8\times$B200 nodes.

The estimand needs replicates, the confounds need ablations, and the boundary needs models outside
the frontier, so the study totals \textbf{2{,}382{,}875 rollouts across 505 cells and 8 models}
(Table~\ref{tab:campaign-main}). Three checks precede every claim: language routing is verified from
each run's own prompt log, where prompt sharing between arms is \textbf{0.0\%}; truncation at 4096
tokens is at most 0.63\% for every model; and coverage is exact, with all 44 arms of the
\emph{audited} campaign holding exactly the planned cells and rollouts
(Appendix~\ref{app:provenance}).

\section{Cross-Lingual Policy Divergence}
\label{sec:divergence}

Table~\ref{tab:convergence} tracks the estimate as each correction is applied. An unmatched
baseline, what an unwary reuse of existing runs would produce, reports $+0.0625$; matching the
budget raises it to $+0.0878$, and strict empty exclusion with a fourth model gives
$\mathbf{+0.0861}$. \textbf{The matched estimate is 38\% larger than the unmatched one.} The worry
is that a confound might be \emph{creating} the effect; here every confound was
\emph{suppressing} it.

\begin{table}[!tb]
\centering
\TableFontSize
\setlength{\tabcolsep}{5.5pt}
\begin{tabular}{@{}lrrrrc@{}}
\toprule
\hdr Estimator & $I_{\text{within}}$ & $I_{\text{cross}}$ & Raw gap & \textbf{Length-matched} & Cells $>0$ \\
\midrule
Unmatched token budget & 0.6634 & 0.5370 & $+0.1264$ & \warncell{$+0.0625$} & 14/15 \\
\rw Matched budget, single-replicate pairing & 0.7103 & 0.6049 & $+0.1054$ & $+0.0878$ & 16/16 \\
Matched replicates, lenient exclusion & 0.6821 & 0.5711 & $+0.1110$ & $+0.0946$ & 18/18 \\
\rw \textbf{Matched replicates, $T{=}0.5$} & 0.7033 & 0.6011 & $+0.1023$ & \keycell{$\mathbf{+0.0861}$} & \textbf{24/24} \\
\textbf{Matched replicates, $T{=}0$} & \textbf{0.8371} & 0.6064 & $+0.2307$ & \keycell{$\mathbf{+0.2074}$} & \textbf{24/24} \\
\bottomrule
\end{tabular}
\caption{Every correction makes the effect larger. Pooled over the four protocol-adherent models,
with GPT-OSS excluded and analysed in \S\ref{sec:correctness}. The uncorrected baseline is shaded
orange and the two headline estimates blue. The last two rows cover identical models and tasks and
differ only in temperature; in both, every 95\% bootstrap interval on the raw gap excludes zero.}
\label{tab:convergence}
\end{table}

If cross-lingual differences were decoding noise correlated with language, removing the noise
should shrink the gap. It does the opposite: on the same four models and tasks the length-matched
gap rises to $\mathbf{+0.2074}$, again positive in 24/24 cells with every bootstrap interval on the
raw gap excluding zero. The decomposition (Figure~\ref{fig:headline}a) is unambiguous: removing
sampling lifts same-language self-consistency sharply and leaves cross-language agreement untouched.
\textbf{Sampling noise was masking the language effect, not producing it}, and $+0.0861$ is a lower
bound on a sampling-free effect of $+0.2074$.

\begin{figure}[!tb]
\centering
\includegraphics[width=\textwidth]{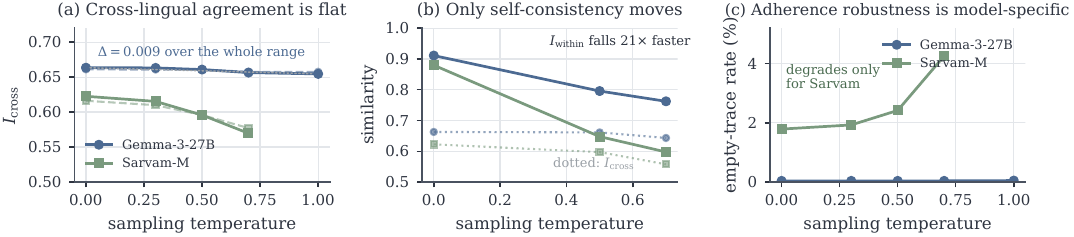}
\caption{The temperature ladder separates language from sampling. (a) Cross-lingual agreement is
flat from greedy to $T{=}1.0$; solid is raw, dashed is length-matched. (b) Self-consistency (solid)
falls steeply while $I_{\text{cross}}$ (dotted) barely moves, so the narrowing gap at high
temperature is a ceiling effect rather than convergence. (c) Adherence under temperature is
model-specific and matters for deployment: Gemma holds a $0.04\%$ empty-trace rate throughout,
whereas Sarvam degrades $2.4\times$ (Appendix~\ref{app:ladder}).}
\label{fig:ladder}
\end{figure}

Two points do not make a curve, so we ran $T \in \{0, 0.3, 0.5, 0.7, 1.0\}$
(Figure~\ref{fig:ladder}). Cross-lingual agreement is essentially flat across the whole range,
while over the sub-range where both are measurable, self-consistency moves twenty-one times
further. This licenses a sharper statement than ``the gap doubles at $T{=}0$'',
which is partly definitional since the gap tends to $1-I_{\text{cross}}$ as decoding becomes
deterministic: \textbf{cross-lingual policy divergence is invariant to sampling temperature}. It is
a property of how the model maps language onto actions rather than of the sampler, and the gap's
apparent temperature-dependence is the ceiling moving, the failure mode C4 predicts. The $T{=}0.7$
arm, paired against the independent five-replicate run, reproduces $\tilde{I}$ to within $0.005$
(Appendix~\ref{app:ladder} against Appendix~\ref{app:voting}).

\begin{figure}[!tb]
\centering
\includegraphics[width=\textwidth]{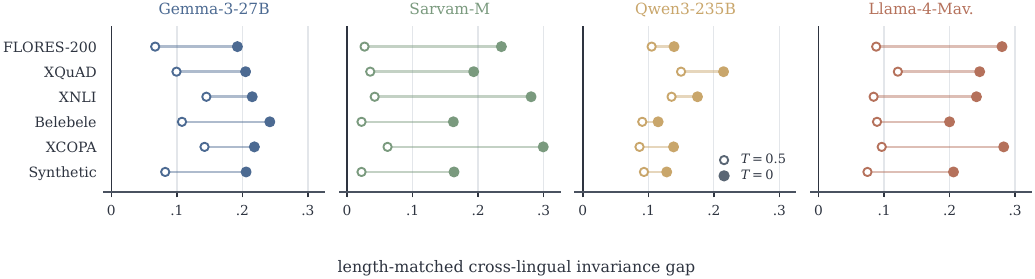}
\caption{Each rule is one (model, benchmark) cell, running from its sampling-inclusive gap (open) to
its greedy gap (filled). All 48 are positive and all 24 rise under greedy decoding, so no single
benchmark or model carries the effect, and at $T{=}0$ every bootstrap interval on the raw gap
excludes zero. Numeric values and intervals are in Appendix~\ref{app:percell}.}
\label{fig:percell}
\end{figure}

The effect is uniform across cells: all 48 length-matched gaps are positive
(Figure~\ref{fig:percell}), including on the synthetic benchmark, whose traces are by far the
longest (Appendix~\ref{app:percell}). Per-model summaries (Table~\ref{tab:permodel}) show the four models
differ substantially in \emph{absolute} gap, precisely the difference \S\ref{sec:estimand} shows is
not what it seems.

\begin{table}[!tb]
\centering
\TableFontSize
\setlength{\tabcolsep}{5pt}
\renewcommand{\arraystretch}{1.08}
\begin{tabular}{@{}lrrrcrrrc@{}}
\toprule
\hdr & \multicolumn{4}{c}{$T = 0.5$ (sampling-inclusive)} & \multicolumn{4}{c}{$T = 0$ (greedy)} \\
\cmidrule(lr){2-5} \cmidrule(lr){6-9}
\hdr Model & $I_{\text{w}}$ & $I_{\text{c}}$ & Len-match & $\tilde{I}$ & $I_{\text{w}}$ & $I_{\text{c}}$ & Len-match & $\tilde{I}$ \\
\midrule
\rw Gemma-3-27B  & 0.7971 & 0.6734 & $+0.1072$ & 0.8311 & 0.9060 & 0.6754 & $+0.2131$ & \keycell{\textbf{0.7276}} \\
Sarvam-M         & 0.6304 & 0.5878 & $+0.0349$ & 0.9219 & 0.8632 & 0.6175 & $+0.2224$ & \keycell{\textbf{0.7076}} \\
\rw Qwen3-235B   & 0.6375 & 0.4977 & $+0.1101$ & 0.7855 & 0.6787 & 0.4902 & $+0.1517$ & \keycell{\textbf{0.7331}} \\
Llama-4-Maverick & 0.7484 & 0.6454 & $+0.0922$ & 0.8553 & 0.9005 & 0.6424 & $+0.2425$ & \keycell{\textbf{0.7092}} \\
\midrule
\hdr \emph{spread} & & & \emph{0.075} & \warncell{\emph{0.136}} & & & \emph{0.091} & \goodcell{\emph{\textbf{0.026}}} \\
\bottomrule
\end{tabular}
\caption{Per-model results at both temperatures. $I_{\text{w}}$ and $I_{\text{c}}$ are
macro-averaged over the six benchmarks whereas $\tilde{I}$ is computed on pooled pairs, so dividing
the two printed columns reproduces $\tilde{I}$ only to within $0.02$; we quote the pooled figure
throughout and reconcile both aggregations in Appendix~\ref{app:percell}. The absolute gap spreads
models widely; the normalised estimand collapses them at $T{=}0$.}
\label{tab:permodel}
\end{table}

\section{Policy Retention and Its Boundary}
\label{sec:estimand}

A model that cannot reproduce itself has no room to display a cross-lingual gap. Across the four
adherent models, the correlation between a model's own self-consistency and its measured gap is
$r = +0.43$ at $T{=}0.5$ and $\mathbf{r = +0.97}$ at $T{=}0$: at greedy decoding, essentially all
between-model variation in the headline gap is explained by the models' own reproducibility rather
than their cross-lingual behaviour. Any ranking built on the absolute gap is therefore a ranking
of determinism wearing a multilingual label, and the consequence is concrete: ordering models by the
$T{=}0.5$ gap and by the $T{=}0$ gap gives rank correlation $\rho = -0.80$
(Table~\ref{tab:percellret}), two temperatures and near-opposite orderings on the same models.
\textbf{No per-model ranking derived from an uncorrected cross-lingual gap should be published.}

Dividing by the ceiling removes the contamination. Under
$\tilde{I} = I_{\text{cross}}/I_{\text{within}}$, bootstrapped directly at the task level, the four
models land within \textbf{2.6 percentage points} of each other at $T{=}0$, every interval narrower
than the band itself (Table~\ref{tab:permodel}, Figure~\ref{fig:headline}b). A 27B dense model, a
24B Indic-specialised dense model and two mixture-of-experts systems, spanning an order of magnitude
in scale and entirely different training mixtures, \textbf{all retain 71--73\% of their own
action-policy self-consistency when the language changes}. The relative spread falls from 16.1\% at $T{=}0.5$ to 3.5\% at $T{=}0$.

\begin{table}[!tb]
\centering
\TableFontSize
\begin{minipage}[t]{0.515\textwidth}
\vspace{0pt}
\centering
\setlength{\tabcolsep}{4pt}
\renewcommand{\arraystretch}{1.07}
\begin{tabular}{@{}lccc@{}}
\toprule
\hdr Grouping & Mean $\tilde{I}$ & Range & Width \\
\midrule
\hdr \multicolumn{4}{@{}l}{\emph{by benchmark}} \\
Belebele & 0.776 & 0.730--0.815 & 0.085 \\
\rw XCOPA & 0.720 & 0.647--0.776 & 0.129 \\
XQuAD & 0.719 & 0.614--0.776 & 0.161 \\
\rw FLORES-200 & 0.713 & 0.675--0.756 & 0.081 \\
XNLI & 0.712 & 0.659--0.771 & 0.112 \\
\rw Synthetic & \keycell{0.691} & 0.676--0.705 & \keycell{0.028} \\
\midrule
\hdr \multicolumn{4}{@{}l}{\emph{by model}} \\
Gemma-3-27B & 0.742 & 0.676--0.772 & 0.095 \\
\rw Qwen3-235B & 0.720 & 0.614--0.794 & 0.180 \\
Llama-4-Maverick & 0.713 & 0.682--0.767 & 0.084 \\
\rw Sarvam-M & 0.713 & 0.647--0.815 & 0.168 \\
\midrule
\textbf{All 24 cells} & \goodcell{\textbf{0.722}} & 0.614--0.815 & SD \goodcell{\textbf{0.052}} \\
\bottomrule
\end{tabular}
\end{minipage}\hfill
\begin{minipage}[t]{0.465\textwidth}
\vspace{0pt}
\centering
\setlength{\tabcolsep}{9pt}
\renewcommand{\arraystretch}{1.07}
\begin{tabular}{@{}lcc@{}}
\toprule
\hdr & $T{=}0$ & $T{=}0.5$ \\
\midrule
\hdr \multicolumn{3}{@{}l}{\emph{variance in $\tilde{I}$ explained by} ($\eta^2$)} \\
Model identity & \goodcell{5.7\%} & \warncell{74.8\%} \\
\rw Benchmark & 26.9\% & 6.9\% \\
Residual & 67.4\% & 18.3\% \\
\midrule
\hdr \multicolumn{3}{@{}l}{\emph{spread across the four models}} \\
Band width in $\tilde{I}$ (pts) & \goodcell{2.6} & \warncell{13.6} \\
\rw Relative spread & 3.5\% & 16.1\% \\
\midrule
\hdr \multicolumn{3}{@{}l}{\emph{rank by uncorrected absolute gap}} \\
Qwen3-235B & \warncell{4} & 1 \\
\rw Gemma-3-27B & 3 & 2 \\
Llama-4-Maverick & \warncell{1} & 3 \\
\rw Sarvam-M & 2 & 4 \\
\bottomrule
\end{tabular}
\end{minipage}
\caption{Policy retention per cell at $T{=}0$ over $n=24$ cells, and what moves it. Left: retention
by benchmark and by model; the ranges overlap heavily, which is the per-cell form of the claim.
Right: one-way $\eta^2$, the width of the across-model band, and the four models ranked by
their uncorrected absolute gap. Under greedy decoding model identity explains almost nothing and the
benchmark dominates; at $T{=}0.5$ that reverses and the ranking inverts with it ($\rho = -0.80$), on
identical models and tasks. Per-cell predictors are in Appendix~\ref{app:predictors}.}
\label{tab:percellret}
\end{table}

The obvious objection is $n=4$, and the data already answers it. Because $\tilde{I}$ is defined per
cell there are 24 of them, and across all 24 it has an SD of just $0.05$. Decomposing that variance
is the informative step. \textbf{Model identity explains 5.7\% of it; the benchmark explains
26.9\%.} That is small in absolute terms, and smaller still against the same estimand on the same
cells under sampling, where model identity explains $74.8\%$: removing the sampler collapses the
model term thirteenfold. The claim is therefore stronger than ``four models agree'':
\textbf{policy retention is close to a model-independent quantity}, and what looks like a per-model
multilingual characteristic at $T{=}0.5$ is largely its own sampling noise. The regularity is
specific to greedy decoding, so any use of the 71--73\% figure must state the temperature.
Correcting for the measured chance floor lowers the level roughly fivefold and preserves the
\emph{absolute} band, though not the relative one, since a fixed spread on a five-times-smaller
base is a five-times-larger relative spread. We report both, and say which the claim rests on
(Appendices~\ref{app:percell} and~\ref{app:chancefloor}).

One model is anomalously irreproducible under greedy decoding, and the natural explanation, expert
routing in mixture-of-experts serving, is refuted by the fourth model, itself an MoE and among the
most reproducible in the study. Trace length is the strongest per-cell predictor, and
\S\ref{sec:mechanism} establishes that it is causal (Appendix~\ref{app:lengthcausal}).

\begin{figure}[!tb]
\centering
\includegraphics[width=\textwidth]{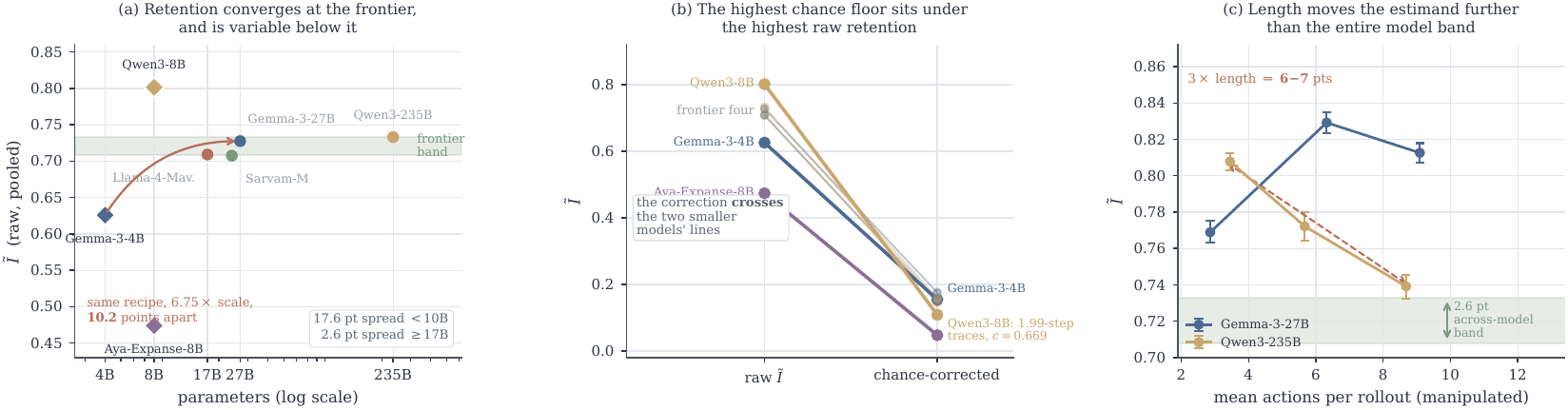}
\caption{The regime, its boundary, and why the boundary is a measurement result. (a) $\tilde{I}$
against parameter count, with the shaded band marking the frontier range; the arrow marks the
within-family comparison of same recipe, $6.75\times$ scale, ten points apart. (b) Raw against
chance-corrected $\tilde{I}$, where the correction crosses the two smaller models' lines. (c) The
three-level length manipulation moves $\tilde{I}$ further than the entire across-model band.}
\label{fig:regime}
\end{figure}

Four models is a striking regularity but a bounded claim, so we extended the study to three systems
chosen so a break would be informative: one shares a family and training recipe with a model already
in the set but is nearly seven times smaller, isolating scale; one shares a family but differs in
training mixture; one comes from an independent vendor, explicitly post-trained for multilinguality. \textbf{At least two of the three fall outside the band, on every way we tried to
compute it} (Figure~\ref{fig:regime}a).

It is not a scaling law. The sharpest line is the within-family comparison: same recipe,
$6.75\times$ the parameters, ten points apart. Nor does scale order the small models: corrected for
chance, the 4B model lands \emph{inside} the frontier band while the 8B falls below it, the opposite
of what a scaling account predicts. What replaces the constant is a regime, retention converging at
the frontier and variable below it with an eightfold difference in SD, offered as a hypothesis at
$n=2$.

This is where measuring the chance floor pays. One 8B model emits the shortest traces in the study
and carries the highest floor measured: two of its traces answering \emph{different} tasks already
score $0.669$, and $0.947$ on one benchmark where its cross-language agreement is $0.9497$.
Corrected, it falls out of the band while the 4B model moves in, so the correction does not
merely rescale the numbers,
it reverses which of the two smaller models looks better (Figure~\ref{fig:regime}b), and no
statement about small-model retention is interpretable on the raw scale.

The obvious alternative is that the frontier four merely occupy a narrow band of trace lengths. One
model refutes this structurally: its mean trace length sits \emph{inside} the frontier range and it
still misses by eight points. Across the six
protocol-adherent models length explains under a third of the variance in retention, so it is a
moderate correlate rather than the explanation. We report the independent-vendor model as an
\emph{adherence} failure rather than a retention datapoint, since it emits no parseable trace on
nearly a third of its rollouts, which leaves the independent-vendor question open. It also produces an
observation worth stating plainly: the two systems in this study specialised for multilinguality are
the two that do worst at multilingual agentic behaviour, one on protocol adherence
(Appendix~\ref{app:scale}) and one on English advantage (Appendix~\ref{app:correctness}).

\section{The English Pivot}
\label{sec:mechanism}

Two observations point the same way. \texttt{Translate} is the \emph{most frequently used tool in
every adapted benchmark}, for every compliant model, while the synthetic benchmark, built around
arithmetic composition, correctly inverts to \texttt{Calc}. And the models' reasoning text is
$\approx$\textbf{99\% ASCII even when the prompt is Devanagari, Tamil or Odia}, although the prompts
are verifiably in-language (\S\ref{sec:setup}). Agents translate into English, plan in English, and
answer: the behavioural counterpart of interpretability results showing that multilingual
transformers pivot through English-aligned latent representations~\citep{wendler2024llamas,
zhao2024multilingualism}, but unlike hidden-state evidence it can be tested by intervention. The
reliance is uneven, with Qwen3 spending \textbf{50.5\%} of its tool calls on \texttt{Translate}
against Sarvam's 18.0\%, which is what makes the mechanism testable rather than merely observable.

\begin{figure}[!tb]
\centering
\includegraphics[width=0.88\textwidth]{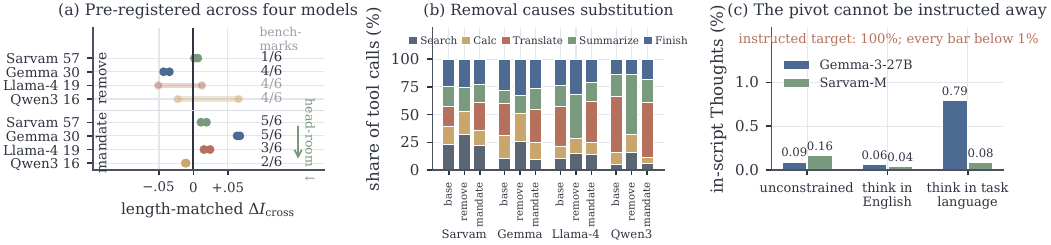}
\caption{English pivoting, tested causally and pre-registered across four models. (a) Length-matched
$\Delta I_{\rm cross}$; two dots per row are the two matching directions and a faded bar means they
disagree; the number beside each model is its head-room and the right column counts benchmarks
signed as predicted. (b) Removal causes substitution, not omission. (c) Instructing models to reason
in the task's language fails, at under 1\% compliance.}
\label{fig:mechanism}
\end{figure}

We manipulate the pivot with two matched arms on \emph{four} models, holding seed, temperature,
budget and task set fixed and changing only the tool alphabet or an ordering constraint:
\texttt{Translate} \emph{removed} from the offered tools, and \texttt{Translate} \emph{mandated} as
the first action on non-English tasks. \textbf{Seven of the eight manipulations took cleanly}:
removal eliminated the tool completely in all four models, none of which reached for a tool it was
not offered (6 calls in ${\sim}500{,}000$), and the mandate took on \textbf{98.7--99.6\%} of
non-English rollouts in three of them; Sarvam-M is the exception at $81.4\%$, flagged wherever its
magnitude is quoted. Removal causes \emph{substitution rather than omission}, with a model-specific
substitute: Gemma into \texttt{Search}/\texttt{Calc}, Qwen3 and Llama-4 into \texttt{Summarize},
Sarvam into \texttt{Search} (Figure~\ref{fig:mechanism}b).

\begin{table}[!tb]
\centering
\TableFontSize
\setlength{\tabcolsep}{3.4pt}
\renewcommand{\arraystretch}{1.10}
\begin{tabular}{@{}lc rrrc rrrrc@{}}
\toprule
\hdr & & \multicolumn{4}{c}{\texttt{Translate} \emph{removed}} & \multicolumn{5}{c}{\texttt{Translate} \emph{mandated}} \\
\cmidrule(lr){3-6} \cmidrule(lr){7-11}
\hdr Model & Head-room & Raw $\Delta$ & A & B & Bench & 1st act. & Raw $\Delta$ & A & B & Bench \\
\midrule
\rw Sarvam-M & \textbf{57.2} & $+0.020$ & $+0.0015$ & $+0.0064$ & 1/6 & \warncell{81.4\%} & $-0.005$ & $+0.0110$ & $+0.0190$ & \goodcell{\textbf{5/6}} \\
Gemma-3-27B & 30.4 & $+0.023$ & \keycell{$\mathbf{-0.0432}$} & \keycell{$\mathbf{-0.0348}$} & 4/6 & 99.6\% & $+0.032$ & \goodcell{$\mathbf{+0.0635}$} & \goodcell{$\mathbf{+0.0674}$} & \goodcell{\textbf{5/6}} \\
\rw Llama-4-Maverick & 18.7 & $+0.055$ & \keycell{$-0.0511$} & \warncell{$+0.0124$} & 4/6 & 98.7\% & $-0.007$ & $+0.0242$ & $+0.0154$ & 3/6 \\
Qwen3-235B & 15.7 & $+0.062$ & \keycell{$-0.0229$} & \warncell{$+0.0659$} & 4/6 & 99.0\% & $+0.073$ & $-0.0105$ & $-0.0117$ & \warncell{2/6} \\
\bottomrule
\end{tabular}
\caption{The English-pivot ablation on all four models, ordered by head-room, which is $100$ minus
the baseline share of non-English rollouts opening with \texttt{Translate}; ``1st act.'' is that
share under the mandate. A and B are the two length-matching directions on $I_{\text{cross}}$ and
``Bench'' counts the benchmarks of six signed as predicted with both agreeing. \textbf{Every removal
arm looks beneficial raw, yet three of the four are harmful once length is matched} (blue), on the
per-benchmark reading the orange cells make necessary. Detail in Appendix~\ref{app:pivot}.}
\label{tab:pivot}
\end{table}

The raw numbers invert the conclusion. Read without length control, \emph{every} arm that
\emph{deletes} the pivot appears to help, on all four models (Table~\ref{tab:pivot}). Removing a
tool shortens traces, and shorter traces score higher similarity, so the raw estimate simply reads
that compression back. Length-matched in both directions (Figure~\ref{fig:mechanism}a), removing
the English pivot lowers cross-lingual agreement in proportion to how much a model uses it: the
effect holds per benchmark for the three models that pivot heavily and not for the one that pivots
least. This is a dose--response refinement of the mechanism rather than a counterexample, and the
substitution pattern agrees: withdrawing the tool sends the freed budget to a different
English-producing operation (Appendix~\ref{app:pivot}).

Our first reconciliation of \emph{why} the mandate helps some models and not others was post-hoc, so
we tested it in advance: two further models were chosen on their measured baseline pivot rate, one
with more head-room than any tested and one with almost none, with both predictions written into the
analysis script before the compute was spent. Across four models the mandate's
benefit is \textbf{monotone in head-room}, at $5/6$, $5/6$, $3/6$ and $2/6$ benchmarks
(Table~\ref{tab:pivot}). Head-room governs \emph{whether} it helps, not how much, since the model
that moved furthest gained $4.4\times$ less than the one that moved second-furthest, so we report
the direction as tested and offer no general prescription.

If English reasoning were a stylistic default, one scaffold line should change it. We tested two
further arms: write every \texttt{Thought:} in English, or in the task's language.
\textbf{The second manipulation was refused.} Told explicitly to reason in the task's language,
Gemma writes a non-Latin-script thought on \textbf{0.79\%} of non-Latin-script rollouts and Sarvam
on \textbf{0.08\%} (Figure~\ref{fig:mechanism}c). We report this as a \emph{failed manipulation},
not a null: the instructed condition was never achieved, so the experiment cannot test whether
reasoning language affects invariance. What it establishes is stronger than the null it replaces:
\textbf{the English pivot is not a preference prompting can switch off}, surviving a direct
instruction in two models at a refusal rate above 99\%. That also explains the asymmetry above: a
scaffold line can only \emph{add} pivoting where it was absent.

Across the adherent cells, mean trace length alone explains two-thirds of the variance in raw
invariance, but that is correlational, so we manipulated it. A three-level intervention holding
task, language, seed and budget fixed moves mean trace length roughly threefold, and moves
$\tilde{I}$ by \textbf{6--7 points with disjoint intervals, further than the entire band across the
four frontier models} (Figure~\ref{fig:regime}c). So the normalisation does \emph{not} absorb
length, and benchmark-level claims resting on it being length-neutral must be withdrawn; length is
not a nuisance to correct away but a first-order driver of agentic policy retention, which is why
every comparison here is length-matched in both directions.
The two models tested disagree on the \emph{sign}, so we make no universal claim about the
direction (Appendix~\ref{app:lengthcausal}).

\section{Outcomes and Measurement Validity}
\label{sec:correctness}

A reasonable objection is that traces are means, not ends: if the answers are right, who cares how
the agent got there? We scored correctness against gold on the four benchmarks that have it, at
14{,}000 gold-bearing rollouts per condition. The answer cuts both ways.

\begin{figure}[!tb]
\centering
\includegraphics[width=0.88\textwidth]{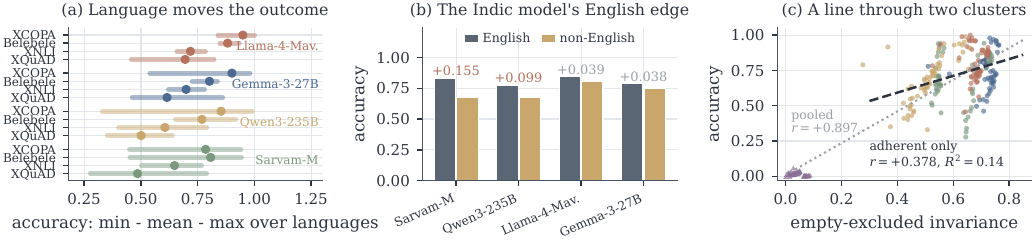}
\caption{Correctness moves with language, but invariance does not predict it. (a) Accuracy spans up
to $0.66$ across languages within one model and benchmark. (b) Every model has an English advantage,
the Indic-specialised model largest. (c) The pooled $r={+}0.897$ is an artifact of two clusters;
excluding GPT-OSS it falls to ${+}0.378$.}
\label{fig:correctness}
\end{figure}

\begin{figure}[!tb]
\centering
\includegraphics[width=0.88\textwidth]{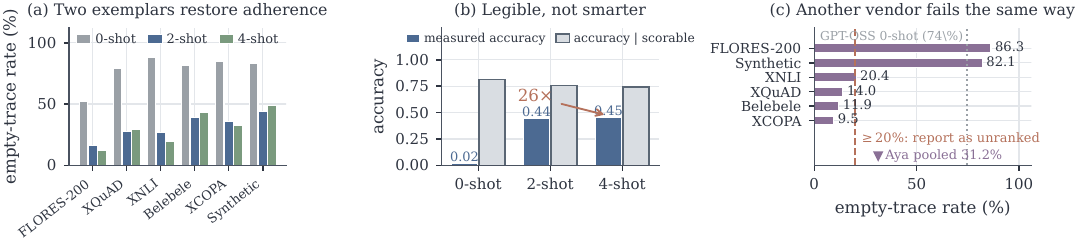}
\caption{Measurement pathology and its repair. (a) GPT-OSS-120B's empty-trace rate under
0/2/4-shot exemplars. (b) Measured accuracy rises $26\times$ while accuracy among scorable rollouts
is flat: legibility, not capability. (c) Aya-Expanse-8B emits no parseable trace on $31.2\%$ of
rollouts. Dashed: $20\%$ unranked threshold; dotted: GPT-OSS 0-shot.}
\label{fig:measurement}
\end{figure}

The language effect does reach the outcome. Cross-lingual accuracy spread reaches \textbf{0.66
within a single model and benchmark}: Qwen3 scores $0.990$ on its best XCOPA language and $0.330$ on
its worst, and Sarvam spans $0.278$ to $0.789$ on XQuAD (Figure~\ref{fig:correctness}a). Averaged
over the four gold benchmarks every model has an English advantage, and the ordering is
counter-intuitive: \textbf{Sarvam-M, the Indic-specialised model, has the largest} at $+0.155$
(Figure~\ref{fig:correctness}b). Specialising a model for a language family does not, on this
evidence, close the English gap in \emph{agentic} performance (Appendix~\ref{app:correctness}).

Invariance is nonetheless not a proxy for accuracy. Pooled across all cells the correlation looks
decisive ($r = +0.897$) but must not be used: it is a line through two clusters, one of them GPT-OSS
measuring a different variable (Figure~\ref{fig:correctness}c). Among adherent models the
association is positive but moderate and \emph{reverses} in a quarter of cells, so invariance and
correctness must be measured separately.

The empty-trace rate that C3 forces us to report separately is itself a finding, collapsing
deterministically for low-resource languages. One model emits no tool call at all on a fifth of the
hardest task family, and the failures are neither truncation nor
noise: the same task in the same language fails in both replicates, on low-resource Indic languages, 22 of 23 affected. This is distinct from policy divergence: the model does not act differently, it fails to
act at all.

The sharpest case is a regex that made a competent model look broken. GPT-OSS-120B yields no
parseable trace on 76.4\% of its rollouts and scores under 2\% accuracy, yet posts the study's two
\emph{highest} raw invariance scores, because empty pairs score $1.0$. The model is not broken; the
parser is. It answers in prose that names the tool, writing ``We will use Translate.''
instead of the required syntax; most non-empty failures are machine-recoverable. Two worked
exemplars, with model, decoding and tasks fixed, raise its measured accuracy
\textbf{twenty-sixfold} while its accuracy on readable outputs barely moves
(Figure~\ref{fig:measurement}): legible, not smarter. Which of the two numbers was ever measuring
capability? Single-pattern extractors are standard in ReAct-style
harnesses~\citep{yao2023react,liu2024agentbench} and parse failure need not be language-uniform, so
the artifact can masquerade as a multilingual finding. We recommend reporting parse-failure rate
with every headline number and treating any model above ${\sim}20\%$ as unranked; a further vendor
fails that threshold at $31.2\%$ (Figure~\ref{fig:measurement}c). The taxonomy, and the finding that
exemplars fix adherence but \emph{not} invariance, are in Appendix~\ref{app:gptoss}.

Inference-time repair does not survive the ceiling either. Self-consistency
voting~\citep{wang2023selfconsistency} is the obvious remedy, and ten replicates let us form
$\tilde{I}$ under voting for the first time, since both sides of the ratio must be voted. Voting
raises raw cross-language agreement in both tested models but same-language agreement more, so on
the ceiling-corrected estimand it costs $1.6$--$1.9$ points with disjoint intervals: a variance
reducer, not a retention improver. Most of our corrections make the effect larger
(Table~\ref{tab:convergence}); this one and the scale extension of \S\ref{sec:estimand} remove
results we could have claimed. We report both, because a correction pipeline that only
ever helps is not one a reader should trust (Appendix~\ref{app:voting}).

\section{Conclusion}
\label{sec:conclusion}

\looseness=-3
Answer agreement is not behavioural agreement. Given the same task in a different language, current
tool-using models take a measurably different route, and under greedy decoding four frontier systems
keep about the same share of their own action policy. Below roughly 10B it breaks, and the ordering
among smaller models is largely an artifact of a chance floor we measured rather than assumed. The
loss is structural, driven by an English pivot the models will not abandon when instructed to and by
a causally established trace-length effect; and much of what looks like multilingual failure is
measurement failure, one regex having suppressed a competent model's accuracy twenty-sixfold.
\textbf{An agent equally accurate in two languages is not thereby equally trustworthy in both},
because cost, failure modes and auditability live in the route rather than the answer, and only an
evaluation that reads the action policy can see the difference.

\bibliography{colm2026_conference}
\bibliographystyle{colm2026_conference}

\clearpage
\appendix
\begin{center}
{\Large\bfseries Appendix}
\end{center}

\vspace{2pt}

\section*{Appendix Guide}

The appendices are grouped by what they are for. Method and estimator come first, then the data and
the checks on it, then the full results behind each figure in the main text, then the boundary of
the regularity, then housekeeping. Within each group the sections run in order.

\begin{center}
\TableFontSize
\TableFontSize
\setlength{\tabcolsep}{5pt}
\renewcommand{\arraystretch}{1.10}
\begin{tabularx}{\textwidth}{@{}l l Y@{}}
\toprule
\hdr & Appendix & What it contains \\
\midrule
\emph{Reference} & \ref{app:keyterms} Key terms & \emph{rollout}, \emph{cell}, \emph{arm}, \emph{confound} and the rest, defined once \\
\midrule
\rw & \ref{app:protocol} Measurement protocol & the estimand, the five confounds, the controls \\
\rw & \ref{app:formal} Estimator details & why the baseline is needed for identification, and the inference \\
\rw & \ref{app:related} Related work & where this sits against agentic and cross-lingual evaluation \\
\rw \multirow{-4}{*}{\emph{Method}} & \ref{app:confounds} The five confounds & each one priced on our own data \\
\midrule
\multirow{3}{*}{\emph{Data}}
 & \ref{app:data} Data and coverage & sources, alignment keys, languages, the synthetic suite \\
 & \ref{app:validity} Harness validity & routing, truncation and coverage, verified in the generated data \\
 & \ref{app:design-checks} Design checks & symmetry, seed, replicate agreement, subset representativeness \\
\midrule
\rw & \ref{app:percell} Per-cell results & all 48 matched cells with intervals \\
\rw & \ref{app:predictors} Trace length & what predicts irreproducibility, and the causal length manipulation \\
\rw & \ref{app:raw} Uncorrected metrics & what a pipeline without the corrections would report \\
\rw & \ref{app:gptoss} GPT-OSS failure taxonomy & the $26\times$ accuracy artifact in full \\
\rw & \ref{app:ladder} Temperature ladder & all five rungs, with adherence \\
\rw & \ref{app:pivot} English-pivot ablation & four models, including the pre-registered head-room test \\
\rw & \ref{app:voting} Self-consistency voting & the ten-replicate result \\
\rw \multirow{-8}{*}{\emph{Results}} & \ref{app:correctness} Correctness & accuracy by language, and why it is not a proxy \\
\midrule
\multirow{2}{*}{\emph{Boundary}}
 & \ref{app:chancefloor} The chance floor & the permutation null, and what correction changes \\
 & \ref{app:scale} Scale and vendor & the three models outside the frontier four \\
\midrule
\rw & \ref{app:limitations} Limitations & what remains open \\
\rw & \ref{app:prompts} Prompts & the scaffold and every intervention \\
\rw & \ref{app:compute} Compute & hardware and cost \\
\rw \multirow{-4}{*}{\emph{Housekeeping}} & \ref{app:provenance} Provenance & the checks that fix which runs the numbers come from \\
\bottomrule
\end{tabularx}
\end{center}

\section{Key Terms}
\label{app:keyterms}

Several terms in Table~\ref{tab:keyterms} name units of the experimental design rather than standard
usage, so they are fixed here once and used consistently throughout the paper.

{\TableFontSize
\setlength{\tabcolsep}{6pt}
\renewcommand{\arraystretch}{1.12}
\begin{xltabular}{\textwidth}{@{}l Y@{}}
\caption{Key terms, fixed here and used consistently throughout the paper.}
\label{tab:keyterms}\\
\toprule
\hdr Term & Meaning in this paper \\
\midrule
\endfirsthead
\toprule
\hdr Term & Meaning in this paper \\
\midrule
\endhead
\bottomrule
\endfoot
\rw \multicolumn{2}{@{}l}{\emph{units of the experimental design}} \\
\textbf{rollout} & One model run on one (task, language) pair. It yields one action trace and one final answer. All counts of the form ``2.38M'' are rollouts. \\
\rw \textbf{trace} & The sequence of tool calls a rollout emits, for example \texttt{Translate}, \texttt{Search}, \texttt{Finish}. This, not the answer, is what we compare. \\
\textbf{action policy} & The mapping a model induces from a task to a trace. Because it is indexed by language, we can ask whether one model has one policy or several. \\
\rw \textbf{cell} & One (model, benchmark, condition) unit, holding all the rollouts for that combination. ``24 cells'' means four models $\times$ six benchmarks. \\
\textbf{arm} & One experimental treatment applied to one model, for instance ``\texttt{Translate} removed, Gemma-3-27B''. An arm spans six cells. \\
\rw \textbf{replicate} & An independent regeneration of a cell under a different serving seed, with everything else held fixed. \\
\midrule
\rw \multicolumn{2}{@{}l}{\emph{the estimator}} \\
$I_{\text{within}}$, $I_{\text{cross}}$ & Same-language and cross-language trace agreement, both measured between the two replicates. \\
\rw \textbf{gap} $\Delta$ & $I_{\text{within}} - I_{\text{cross}}$: the raw shortfall, before any normalisation. \\
$\tilde{I}$ & Normalised policy retention, $I_{\text{cross}}/I_{\text{within}}$: the share of a model's own reproducibility that survives a change of language. \\
\rw \textbf{ceiling-corrected} & Expressed as $\tilde{I}$ rather than as $\Delta$, so that a model's own irreproducibility is divided out instead of being read as a language effect. \\
\textbf{frontier band} & The $[0.708, 0.733]$ range of $\tilde{I}$ that the four frontier models occupy at $T{=}0$; ``outside the band'' is measured against it. \\
\midrule
\rw \multicolumn{2}{@{}l}{\emph{corrections and their objects}} \\
\textbf{confound} & A property of the \emph{measurement} rather than of the models that moves the reported number by as much as the effect being measured. We price five (Appendix~\ref{app:confounds}). \\
\rw \textbf{length matching} & Reweighting the pairs being compared to a common trace-length distribution before averaging, done in both directions and accepted only when both agree in sign. \\
\textbf{chance floor $c$} & What the similarity score $S$ returns for two traces that answer \emph{different} tasks. Measured by permutation, not assumed (Appendix~\ref{app:chancefloor}). \\
\rw \textbf{adherence} & Whether a model emits a parseable trace at all. A model that does not is reported but excluded from retention comparisons. \\
\textbf{parse failure} & A rollout whose output contains no extractable \texttt{Action:} call. A subset are \emph{empty traces}, where the output is blank; the rest name a tool in prose. \\
\midrule
\rw \multicolumn{2}{@{}l}{\emph{the mechanism}} \\
\textbf{English pivot} & The habit of translating a non-English task into English before planning in it, visible as \texttt{Translate} opening a trace and as English-language reasoning text. \\
\rw \textbf{load-bearing} & Said of the pivot: removing it degrades cross-lingual agreement, so it is doing causal work rather than being an incidental stylistic habit. \\
\textbf{head-room} & $100$ minus a model's baseline rate of opening a non-English rollout with \texttt{Translate}: how much room an instruction to pivot has to change behaviour. \\
\rw \textbf{pre-registered} & Of a prediction: written into the analysis script, with the models chosen on a measured baseline property, before the compute was spent. \\
\end{xltabular}
}

\section{Measurement Protocol}
\label{app:protocol}

Let $z$ be a \emph{canonical task} and $x^{(\ell)}(z)$ its realisation in language $\ell$, taken
from a verified parallel alignment key rather than a row index (\S\ref{sec:setup}). Under a fixed
scaffold with tool alphabet $\mathcal{T} = \{\texttt{Search}, \texttt{Calc}, \texttt{Translate},
\texttt{Summarize}, \texttt{Finish}\}$, a model emits an executed action trace
$A^{(\ell)}(z) \in \mathcal{T}^{*}$ alongside its answer; the trace, not the answer, is what we
compare. With $S(A,B) \in [0,1]$ the normalised matching-block similarity, the naive corpus
estimate is $I_{\text{cross}} = \mathbb{E}_{z,\ell_i \neq \ell_j} S(A^{(\ell_i)}, A^{(\ell_j)})$
(Appendix~\ref{app:formal}).

That quantity has no meaningful zero, and its maximum is attainable by a degenerate policy: a
model that always emits \texttt{Finish} scores $1.0$. Worse, four confounds move it by as much as
the effect being measured: a missing baseline, trace length, empty traces, and the reproducibility
ceiling. Table~\ref{tab:confounds} quantifies all four on this paper's own data, and three of them
reverse a sign or a ranking there. A fifth, which we had to measure rather than correct, is the
subject of the next paragraph.

The metric has no zero, so we measure it. $S$ is a sequence-similarity score, and with a five-tool alphabet and short traces two
\emph{unrelated} policies already agree well above zero. Because $\tilde{I}$ is a ratio it is
strictly more sensitive to an uncorrected floor than the difference is: under the standard model
$S_{\text{obs}} = c + (1-c)\,S_{\text{true}}$ the difference rescales cleanly and its ordering is
invariant to $c$, whereas the ratio is biased toward $1$ and differentially so by model. We
therefore \emph{measure} $c$ rather than assume it, by permuting the task\,$\rightarrow$\,trace
assignment within a language arm and within length bin and recomputing $S$: the exact null
``what does $S$ score when two traces answer \emph{different} tasks?'' Across \textbf{66 cells}
and 200 permutations, $\mathbf{c \approx 0.56}$, ranging from $0.35$ on the longest-trace
benchmark to $\mathbf{0.95}$ on the shortest. This is the difference between a hedge and a
measurement, and \S\ref{sec:estimand} shows that it decides a headline ranking
(Appendix~\ref{app:chancefloor}).

We generate every cell \textbf{twice} under identical decoding configuration, task set and token
budget, varying only the serving seed. Writing $r_1, r_2$ for the replicates,
\begin{equation}
I_{\text{within}} = \mathbb{E}_{z,\ell}\, S\big(A^{(\ell)}_{r_1}, A^{(\ell)}_{r_2}\big),
\qquad
I_{\text{cross}} = \mathbb{E}_{z,\ell_i \neq \ell_j}\,
S\big(A^{(\ell_i)}_{r_1}, A^{(\ell_j)}_{r_2}\big).
\end{equation}
Both sides are \emph{cross-seed}, so decoding noise enters identically and language identity is
the only difference. The \textbf{gap} is $\Delta = I_{\text{within}} - I_{\text{cross}}$ and
\textbf{normalised policy retention} is $\tilde{I} = I_{\text{cross}}/I_{\text{within}}$: the
share of a model's own reproducibility that survives a change of language.

Because $S$ is length-sensitive (C2), every arm-to-arm comparison is reweighted to a common
distribution over $\max(|A|,|B|)$ bins in both directions, and we accept a result only when both
agree in sign. Intervals are task-level bootstraps with 1{,}000
resamples~\citep{efron1993bootstrap}. Under C3 a pair is dropped whenever either trace is empty,
which makes empty-trace and parse-failure rates first-class metrics and a finding in their own
right (\S\ref{sec:correctness}). Design checks are in Appendix~\ref{app:design-checks}.

\section{Estimator Details}
\label{app:formal}

Appendix~\ref{app:protocol} states the estimator; this appendix supplies the properties it relies
on, the identification argument, and the inference details.

Under a fixed scaffold with tool alphabet $\mathcal{T}$, a model
$\theta$ induces $F_\theta(z,\ell) = (A^{(\ell)}(z),\, y^{(\ell)}(z))$, where
$A^{(\ell)}(z) = \langle \tau_1,\dots,\tau_k\rangle \in \mathcal{T}^{*}$ is the executed action
trace and $y^{(\ell)}(z)$ the final answer. Outcome-scored evaluation observes only $y$; we
observe $A$. We treat the induced behaviour as a family of language-indexed policies
$\{\pi_\theta^{(\ell)}\}$~\citep{puterman1994mdp} and ask how far apart its members are. Because
$\mathcal{T}$ is fixed and language-independent, traces from two languages are sequences over the
same alphabet and can be compared exactly, without a translation or entailment model in the
measurement path.

The estimator has to be cross-seed on both sides. $I_{\text{within}}$ and
$I_{\text{cross}}$ are both computed between replicate $r_1$ and replicate $r_2$, never within a
single run. This symmetry is what a single-run estimate cannot offer: if $I_{\text{cross}}$ were
measured across languages within one run while $I_{\text{within}}$ compared two runs, the two
quantities would differ in \emph{two} respects (language and seed) rather than one, and the
difference could not be attributed to language. Under the cross-seed construction decoding noise
enters both sides identically and language identity is the only remaining difference.

The similarity $S(A,B)$ is the normalised matching-block similarity
(the ratio $2M/(|A|+|B|)$, where $M$ is the total size of matched blocks under a longest-
contiguous-match decomposition). It is symmetric, equals $1$ for identical sequences and $0$ for
sequences sharing no tool. Two properties drive the confounds. It has \textbf{no meaningful zero}:
with $|\mathcal{T}|=5$ and short traces, two unrelated policies already score well above zero.
And it is \textbf{length-sensitive}: for a fixed edit rate, shorter sequences score higher, which
is why every arm-to-arm comparison in this paper is length-matched (C2).

$I_{\text{within}}$ is not $\approx 1$, and without it the quantity is not identifiable. A model re-run on the \emph{same} prompt in the \emph{same} language under a
different serving seed does not reproduce its own trace: we measure
$I_{\text{within}} = 0.63$--$0.80$ depending on model and temperature. This is the identification
problem in one line. A cross-language observation $S(A^{(\ell_i)}, A^{(\ell_j)})$ is depressed by
two sources at once, the model's own stochasticity and the change of language, and a single
run supplies one equation in two unknowns. Any number computed from it is a language effect only
under the assumption $I_{\text{within}} = 1$, which is false by $20$--$37$ points here and false
by a \emph{model-dependent} amount, so the assumption does not even preserve orderings
(\S\ref{sec:estimand}). The matched replicate supplies the second equation, and it must be
measured rather than borrowed: reusing an existing run introduces a token-budget mismatch that
alone moves the headline from $+0.0861$ to $+0.0625$, because a smaller budget truncates traces,
which shortens them, which inflates $S$ on both sides but not equally.

Intervals are task-level bootstraps with 1{,}000
resamples~\citep{efron1993bootstrap}. Resampling is over \emph{tasks} rather than over pairs,
because the $\binom{L}{2}$ language pairs generated by one task are not independent; resampling
pairs would understate the intervals substantially. The same resample indices are used for
$I_{\text{within}}$ and $I_{\text{cross}}$ within a draw, so $\tilde{I}$ and $\Delta$ inherit the
correlation between numerator and denominator rather than treating them as independent.

Under C3 a pair is dropped whenever \emph{either} trace is
empty. The strict rule matters: retaining empty-vs-empty pairs is what lifts GPT-OSS-120B to the
top of the raw invariance ranking (Appendix~\ref{app:gptoss}), and the correction is worth up to
$\mathbf{-0.119}$ on a single cell. Because exclusion discards information non-uniformly across
languages, empty-trace and parse-failure rates are reported as first-class metrics everywhere
rather than being silently absorbed into the invariance number.

The symmetry correction, a same-seed control and subset
representativeness are verified in Appendix~\ref{app:design-checks}.

\section{Extended Related Work}
\label{app:related}

ReAct-style scaffolds~\citep{yao2023react}
and learned tool use~\citep{schick2023toolformer, parisi2022talm, qin2024toolllm} established the
interleaved reason--act format we adopt: the model alternates free-form deliberation with a
symbolic call drawn from a declared tool inventory. Agentic benchmarks evaluate task
completion~\citep{liu2024agentbench, mialon2024gaia} but are overwhelmingly English and score
\emph{outcomes}, using the action sequence only as a means to that score. Two consequences follow
for our purposes. First, a model that reaches the right answer by a different route is
indistinguishable from one that reaches it by the intended route, so behavioural variation is
invisible by construction. Second, because scoring depends on extracting a final answer, these
harnesses inherit a parsing step whose failure modes are not audited; \S\ref{sec:correctness} and
Appendix~\ref{app:gptoss} show that this step alone can suppress a model's measured accuracy by
more than an order of magnitude. We instead make the trace the measured object and ask a
comparative question about it across languages, which requires a same-language baseline that
outcome-scored suites do not need and do not collect.

Cross-lingual
suites~\citep{hu2020xtreme, ahuja2023mega, liang2023helm} compare final predictions, and the
underlying datasets we build on~\citep{xnli2018, xcopa2020, xquad2020, belebele2024} are
translations or careful adaptations of a common item pool. Their
alignment guarantees are exactly what makes a matched cross-lingual comparison possible, but they
are guarantees about \emph{items}, not about pipelines: Appendix~\ref{app:data} documents five
multilingual corpora whose per-language splits do not correspond item-for-item despite being
distributed together, and one (Belebele) that requires re-keying before an index join is valid.

Chain-of-thought work and its multilingual extension~\citep{wei2022chain, shi2023multilingualcot}
and multitask finetuning~\citep{muennighoff2023crosslingual} examine intermediate \emph{text}.
Text is the natural object when the question is whether a model reasons at all, but it is a poor
object for a cross-lingual comparison: two rationales in different languages cannot be scored for
agreement without a translation or entailment model, which introduces its own multilingual error
profile into the measurement. Our shared symbolic action alphabet avoids this entirely. The
comparison is between sequences over a fixed, language-independent vocabulary of tool names, so
agreement is computed exactly rather than estimated.

Closest in spirit,
\citet{qi2023crosslingual} measure whether models retrieve the same \emph{facts} across languages
and report substantial inconsistency. We ask the analogous question about \emph{procedures}. The
distinction matters because a procedure is chosen rather than recalled: the model can reach the
same answer through different tools, so procedural divergence is not simply knowledge divergence
observed at another layer. It also changes what must be controlled. A fact is either retrieved or
not, whereas a procedure is sampled, so any cross-lingual procedural comparison confounds language
with decoding noise unless a matched same-language baseline is measured under identical
conditions. Appendix~\ref{app:formal} shows that without this baseline the quantity is not
identifiable, and \S\ref{sec:estimand} shows that the uncorrected version ranks models by their
determinism rather than by their cross-lingual behaviour.

\citet{wendler2024llamas} and
\citet{zhao2024multilingualism} give representational evidence that multilingual transformers
route non-English inputs through English-aligned internal states. That evidence is correlational
with respect to behaviour: it establishes that an English-like representation is present, not that
downstream behaviour depends on it. Our ablation is the behavioural, causal counterpart, removing
the pivot from the action space and observing the effect on cross-lingual agreement
(\S\ref{sec:mechanism}, Appendix~\ref{app:pivot}). The two directions of evidence agree, and the
behavioural version adds a result the representational version cannot supply: the pivot survives a
direct instruction not to use it, at a $>$99\% refusal rate in both models tested.

\citet{sclar2024quantifying} show that
input-formatting choices move benchmark scores substantially, and that this sensitivity is large
enough to reorder models. \S\ref{sec:correctness} documents the output-side analogue: a
single-pattern trace extractor, applied uniformly to all models and languages, suppressed one
model's measured accuracy by $26\times$ and simultaneously promoted it to the top of the raw
invariance ranking. The output-side case is arguably the more dangerous of the two, because a
formatting choice is visible in the prompt whereas an extractor lives in the harness, and because
parse failure need not be language-uniform, so the artifact can present as a multilingual finding.

\section{The Five Confounds}
\label{app:confounds}

Every result in the paper corrects for all five confounds below, and each is large enough on its
own to change a published conclusion. Four do more than that on this paper's own data: C2
\emph{reverses the sign} of two interventions, C3 and C4 \emph{reverse a model ranking} (C3 by
demoting the model whose empty pairs score $1.0$, C4 by ranking determinism rather than language,
$\rho = -0.80$ between two temperatures), and C5 \emph{reverses} which of two smaller models
retains more. C1's cost is a $38\%$ underestimate of the headline rather than a reversal, which is
precisely why it went unnoticed. Impacts are measured on this paper's own data. Two of the five are
established \emph{causally} rather than by correction alone: C2 by a three-level manipulation
(Appendix~\ref{app:lengthcausal}) and C5 by a permutation null over 66 cells
(Appendix~\ref{app:chancefloor}). \textbf{C5, the chance floor:}
$S$ has no meaningful zero, because unrelated policies over a five-tool alphabet already agree
over half the time; measured $c \approx 0.56$ and up to $0.95$ on the shortest traces, and
correcting for it \emph{reverses} which of two smaller models shows higher retention.

\begin{table}[h]
\centering
\TableFontSize
\setlength{\tabcolsep}{3.5pt}
\renewcommand{\arraystretch}{1.12}
\begin{tabularx}{\textwidth}{@{}p{1.35cm} X X p{4.05cm}@{}}
\toprule
\hdr \textbf{Confound} & \textbf{Symptom if ignored} & \textbf{Correction} & \textbf{Measured impact} \\
\midrule
\rw \textbf{C1}\newline no baseline &
Decoding noise is scored as a language effect &
Generate every cell twice; identical budget, seed is the only difference &
\keycell{$I_{\text{within}}\!=\!0.63$--$0.80$, not $\approx\!1$; matching the budget makes the estimate \textbf{38\% larger}} \\
\textbf{C2}\newline trace length &
Short traces score higher, so ``hard benchmark'' and ``long trace'' look alike &
Reweight to a common max-length bin distribution, in \emph{both} directions &
\keycell{$R^2\!=\!0.67$ across cells (0.92 within one model); flips the \emph{sign} of two interventions} \\
\rw \textbf{C3}\newline empty traces &
Two empty traces are maximally similar, so non-adherence is rewarded &
Drop a pair when \emph{either} side is empty; report the empty rate separately &
\keycell{Up to $\mathbf{-0.119}$ on one cell; collapses one model's headline invariance $0.79\!\rightarrow\!0.02$} \\
\textbf{C4}\newline ceiling &
The gap is bounded by a model's own reproducibility, so rankings rank determinism &
Normalise: $\tilde{I} = I_{\text{cross}} / I_{\text{within}}$ &
\keycell{$r(I_{\text{within}}, \text{gap})\!=\!\mathbf{+0.97}$ at $T\!=\!0$; cross-model spread $81\%\!\rightarrow\!34\%$} \\
\rw \textbf{C5}\newline chance floor &
Unrelated traces already agree by construction, so $S$ has no meaningful zero &
Measure the floor by permuting task$\rightarrow$trace within a language arm, within length bin &
\keycell{$c\!\approx\!\mathbf{0.56}$, up to $0.95$ on short traces; \emph{reverses} which of two smaller models retains more} \\
\bottomrule
\end{tabularx}
\caption{The five confounds in cross-lingual policy measurement, each quantified on this paper's
own data. C2 and C5 are additionally established causally, by manipulation and by permutation null
respectively.}
\label{tab:confounds}
\end{table}

C1 dominates the others, and the reason is structural: they are metric artifacts, whereas C1 is
an identification failure. Without a same-language baseline, $I_{\text{cross}}$ has no reference point at all, and
the natural fallback of reusing an existing single run silently introduces a token-budget
mismatch. On our data that mismatch alone moves the headline from $+0.0861$ to $+0.0625$, and it
moves it in the \emph{conservative} direction, which is why the confound went unnoticed until a
matched baseline was generated.

The corrections are not commutative in magnitude, though they are in sign. We apply C1 (matched replicates) at generation time, then C3 (strict empty
exclusion) at pair construction, then C2 (length matching) at aggregation, then C4 (normalisation)
at reporting. Applying C2 before C3 changes the pooled estimate by $<0.004$ because empty pairs
sit in the shortest length bin and are dropped either way.

\section{Data and Coverage}
\label{app:data}

This appendix documents where the tasks come from, what guarantees their cross-lingual
correspondence, and what had to be rebuilt before they could be used. The organising constraint is
stated in Appendix~\ref{app:formal}: a cross-lingual comparison of \emph{procedures} is only
well-posed if $x^{(\ell_i)}(z)$ and $x^{(\ell_j)}(z)$ are the same underlying task $z$. That is a
property of the alignment key, not of the corpus name, so we record the key used for every
benchmark below and treat any dataset without one as unusable rather than as noisy.

\begin{table}[h]
\centering
\TableFontSize
\setlength{\tabcolsep}{5pt}
\renewcommand{\arraystretch}{1.06}
\begin{tabular}{@{}llrrrrl@{}}
\toprule
\hdr Benchmark & Source & Tasks & Langs & Rollouts & Pairs & Alignment key \\
\midrule
\rw FLORES-200 & \citet{nllb2022} & 997 & 21 & 20{,}937 & 209{,}370 & sentence \texttt{id} \\
XQuAD & \citet{xquad2020} & 1{,}189 & 12 & 14{,}268 & 78{,}474 & question \texttt{id} \\
\rw XNLI & \citet{xnli2018} & 2{,}490 & 15 & 37{,}350 & 261{,}450 & \texttt{all\_languages} config \\
Belebele & \citet{belebele2024} & 900 & 16 & 14{,}400 & 108{,}000 & \warncell{\texttt{link}+\texttt{question\_no.}} \\
\rw XCOPA & \citet{xcopa2020} & 100 & 11 & 1{,}100 & 5{,}500 & \texttt{idx} \\
Synthetic & this work & 100 & 23 & 2{,}300 & 25{,}300 & construction \\
\midrule
\hdr \textbf{Total} & & \textbf{5{,}776} & \keycell{\textbf{41}$^\dagger$} & \keycell{\textbf{90{,}355}} & \keycell{\textbf{688{,}094}} & \\
\bottomrule
\end{tabular}
\caption{Evaluation suite, \emph{per model}. $^\dagger$41 distinct languages across the suite,
not a sum. Gold answers exist for XQuAD, XNLI, Belebele and XCOPA; FLORES-200 has none by
construction. XCOPA is the only benchmark with no English arm. The orange cell is the one key we
had to construct ourselves: Belebele ships no usable index, and joining on row order matches only
51\% of items.}
\label{tab:data}
\end{table}

\subsection{The experimental campaign}
\label{app:campaign}

Table~\ref{tab:campaign-main} lists every condition generated for this paper in one place, so that
the scale claim in \S\ref{sec:setup} can be audited row by row rather than taken on trust. Rows are
ordered as the paper uses them: the main sweep and the matched replicates that make the estimand
identifiable, then the conditions that remove sampling, then the six interventions, then the
CPU-only permutation null. Every row is generated under the same harness and the same decoding
configuration except where the row is itself the manipulation.

\begin{table}[!t]
\centering
\TableFontSize
\setlength{\tabcolsep}{5pt}
\renewcommand{\arraystretch}{1.06}
\begin{tabular}{@{}llrrr@{}}
\toprule
\hdr Condition & Purpose & Models & Cells & Rollouts \\
\midrule
\rw Main sweep ($T{=}0.5$) & Raw metrics, adherence & 5 & 30 & 451{,}775 \\
Matched replicates ($T{=}0.5$) & Baseline $I_{\text{within}}$ (C1) & 5 & 30 & 226{,}000 \\
\rw Greedy ($T{=}0$) & Remove sampling noise & 4 & 24 & 180{,}800 \\
Temperature ladder & $T \in \{0, 0.3, 0.5, 0.7, 1.0\}$ & 2 & 30 & 113{,}000 \\
\rw Few-shot exemplars & Adherence repair & 2 & 24 & 90{,}400 \\
Tool-alphabet ablation & Pivot mechanism (causal) & 2 & 24 & 90{,}400 \\
\rw Reasoning-language control & Pivot mechanism (causal) & 2 & 19 & 74{,}100 \\
Pivot head-room \emph{(pre-registered)} & Mechanism, 4 models & 2 & 36 & 135{,}600 \\
\rw Voting, 5 replicates & Inference-time repair & 2 & 60 & 162{,}000 \\
Voting, 10 replicates ($k\!\leq\!5$) & Ceiling-corrected repair & 2 & 120 & 452{,}000 \\
\rw Scale \& vendor generalisation & Boundary of the regime & 3 & 36 & 135{,}600 \\
Trace-length dose--response & C2 causality & 2 & 72 & 271{,}200 \\
\rw Chance floor (permutation) & C5, CPU-only & 8 & \emph{(66)}$^\dagger$ & --- \\
\midrule
\hdr \textbf{Total} & & \textbf{8} & \goodcell{\textbf{505}} & \goodcell{\textbf{2{,}382{,}875}} \\
\bottomrule
\end{tabular}
\caption{The experimental campaign. A \emph{cell} is one (model, benchmark, condition) unit. $^\dagger$The chance-floor analysis is a permutation null recomputed over cells that the rows
above already generated, so it contributes neither rollouts nor cells of its own; its 66 are
parenthesised and excluded from both totals, and the twelve generated rows sum exactly to $505$
and $2{,}382{,}875$. Ablation arms use a 300-task subset per benchmark, which reproduces
full-sweep invariance to within $\pm0.02$ (Appendix~\ref{app:design-checks}).}
\label{tab:campaign-main}
\end{table}

\subsection{Item-level parallelism}

Item-level parallelism is a property of the alignment key, not of the corpus name, and it is
scarcer than it looks. Screening candidate corpora against it rules out Samanantar~\citep{samanantar2022}
and OPUS-100~\citep{opus1002020}, which are En--X bitext with no X--Y correspondence;
XL-Sum~\citep{xlsum2021}, whose articles are collected independently per language;
MLQA~\citep{mlqa2020}, which is parallel only through a shared question id and whose 7 languages
are a strict subset of XQuAD's 12; and TyDi QA~\citep{tydiqa2020}, which is not parallel by design.
A pipeline that indexes any of them by row runs without error and returns a cross-lingual
invariance number that means nothing, which is the failure mode this appendix exists to rule out
for the six benchmarks we do use.

\subsection{Building the suite}

The six benchmarks were not usable as distributed either. Each defect below is an ordinary
data-engineering fault rather than anything exotic, which is precisely why it is easy to inherit
silently: every one yields a corpus that loads, iterates and scores without raising an error.

\begin{table}[h]
\centering
\TableFontSize
\setlength{\tabcolsep}{6pt}
\renewcommand{\arraystretch}{1.18}
\begin{tabularx}{\textwidth}{@{}p{4.3cm} Y p{3.5cm}@{}}
\toprule
\hdr Issue in the distributed data & What we did & Effect \\
\midrule
\rw Belebele carries no usable cross-language index & Re-keyed on \texttt{link}\,+\,\texttt{question\_no.}; a row-order join matches 51\% of items, with gold agreeing on 64\% & \keycell{900/900 items aligned, gold 100\%} \\
Only a subset of each corpus's languages is exposed by default & Rebuilt every benchmark at full language coverage & XNLI 3$\to$15, XQuAD 3$\to$12, FLORES 12$\to$21, Belebele 11$\to$16 \\
\rw Gold answers are not carried through to the task files & Gold emitted at build time & 4 benchmarks, 4{,}679 tasks \\
Some synthetic instances came back untranslated & Regenerated & \keycell{515 of 2{,}200 (23.4\%) $\to$ 0} \\
\rw Some synthetic instances came back romanised & Regenerated in native script & 678 instances, 97.5\% compliant \\
\bottomrule
\end{tabularx}
\caption{What the suite required before any measurement. The two blue cells matter most: a 51\%
item-match rate silently compares different questions across languages, and untranslated instances
make a ``non-English'' arm partly English.}
\end{table}

\subsection{Languages}

Forty-one languages, 17 families and 16 writing systems. Summed over benchmarks the suite yields
809 (benchmark, language-pair) comparisons, over \textbf{428 distinct language pairs}: the
per-benchmark language sets overlap heavily, and Belebele's 16 languages are a subset of
FLORES-200's 21, so Belebele contributes no pair FLORES does not already cover. Resource tier
follows the standard high/medium/low split (23/11/7). The suite is deliberately unbalanced toward
Indic languages, which is what makes the adherence-collapse result of \S\ref{sec:correctness}
visible: a suite of only high-resource European languages would not have exposed it.

\begin{table}[h]
\centering
\TableFontSize
\setlength{\tabcolsep}{3pt}
\renewcommand{\arraystretch}{1.04}
\begin{tabular}{@{}llll|llll@{}}
\toprule
\hdr Code & Language & Family & Script & Code & Language & Family & Script \\
\midrule
\rw ar & Arabic & Semitic & Arabic & mni & Manipuri & Sino-Tibetan & Bengali \\
as & Assamese & Indo-Aryan & Bengali & mr & Marathi & Indo-Aryan & Devanagari \\
\rw bg & Bulgarian & Slavic & Cyrillic & ne & Nepali & Indo-Aryan & Devanagari \\
bn & Bengali & Indo-Aryan & Bengali & or & Odia & Indo-Aryan & Oriya \\
\rw brx & Bodo & Sino-Tibetan & Devanagari & pa & Punjabi & Indo-Aryan & Gurmukhi \\
de & German & Germanic & Latin & qu & Quechua & Quechuan & Latin \\
\rw doi & Dogri & Indo-Aryan & Devanagari & ro & Romanian & Romance & Latin \\
el & Greek & Hellenic & Greek & ru & Russian & Slavic & Cyrillic \\
\rw en & English & Germanic & Latin & sa & Sanskrit & Indo-Aryan & Devanagari \\
es & Spanish & Romance & Latin & sat & Santali & Austroasiatic & Ol Chiki \\
\rw et & Estonian & Uralic & Latin & sd & Sindhi & Indo-Aryan & Arabic \\
fr & French & Romance & Latin & sw & Swahili & Bantu & Latin \\
\rw gom & Konkani & Indo-Aryan & Devanagari & ta & Tamil & Dravidian & Tamil \\
gu & Gujarati & Indo-Aryan & Gujarati & te & Telugu & Dravidian & Telugu \\
\rw hi & Hindi & Indo-Aryan & Devanagari & th & Thai & Kra-Dai & Thai \\
ht & Haitian & French Creole & Latin & tr & Turkish & Turkic & Latin \\
\rw id & Indonesian & Austronesian & Latin & ur & Urdu & Indo-Aryan & Arabic \\
it & Italian & Romance & Latin & vi & Vietnamese & Austroasiatic & Latin \\
\rw kn & Kannada & Dravidian & Kannada & zh & Chinese & Sinitic & Han \\
ks & Kashmiri & Indo-Aryan & Arabic & mai & Maithili & Indo-Aryan & Devanagari \\
\rw ml & Malayalam & Dravidian & Malayalam & & & & \\
\bottomrule
\end{tabular}
\end{table}

Bodo, Dogri and Konkani appear only in the synthetic benchmark; no parallel benchmark reaches
them. Script compliance of the benchmark files is 97.1--100\% per language, measured by Unicode
block membership.

\subsection{Synthetic benchmark composition}

The suite was generated and translated with \texttt{gpt-5-mini} (deployment \texttt{2025-08-07})
at its default decoding settings; the generation prompt is in Appendix~\ref{app:prompts}. No model
evaluated in this paper was used to author it. 100 tasks $\times$ 23 languages. Task families:
constraint planning (18), retrieval--calculation
pipeline (17), evidence reconciliation (17), policy-compliance exception (16),
schedule/resource allocation (16), error-audit correction (16). Difficulty: medium (51), hard
(30), easy (19). Each task carries \texttt{required\_tools}, \texttt{expected\_min\_steps} and
\texttt{complexity\_tags}. Median prompt length is 3{,}233 characters, against 139--800 for the
adapted benchmarks, which is why its traces are $1.7$--$4.6\times$ longer and why it must never
be compared to the adapted benchmarks without length matching.

\section{Harness Validity}
\label{app:validity}

Three properties are verified in the delivered data rather than assumed, because each can fail
silently: a harness that routes every language arm to the same prompt, truncates long traces, or
drops rollouts from a subset of arms will still produce a complete-looking results table. The
checks below are run on the generated data, not on a separate test fixture.

Language routing is measured from each run's own prompt log on Belebele, over 16 arms and 900
tasks. Every arm holds 900 distinct prompts, \textbf{prompt sharing between arms is 0.0\%} for
every model, and mean ASCII is 0.348 with exactly one Latin-script arm, English. A run that routes
every arm to the same text shows 100\% sharing at 0.999 ASCII on every arm; ours shows none.

Truncation, meaning \texttt{finish\_reason == length} at \texttt{max\_tokens}$=$4096, is 0.003\%
for Gemma, 0.005\% for Sarvam, 0.004\% for Llama-4, 0.61\% for Qwen3 and 0.63\% for GPT-OSS, so no
result in this paper is budget-limited. The field \texttt{used\_reasoning\_channel} is $0$ on all
451{,}775 main-sweep rollouts for every model, which excludes by direct measurement rather than by
argument the hypothesis that GPT-OSS's short outputs reflect a separate reasoning channel the
harness discarded (\S\ref{sec:correctness}). And coverage is exact: every cell matches
$\text{tasks} \times \text{languages}$, for 451{,}775 of 451{,}775 rollouts, with no ragged
language arm and no missing task.

\section{Design Checks}
\label{app:design-checks}

The symmetry correction is not an artifact. Our estimator takes $I_{\text{cross}}$
across replicates. Compared against the same-language-run convention, the two agree to within
$0.002$ on every model (Gemma $0.6728$/$0.6734$, Sarvam $0.5855$/$0.5878$, Qwen3
$0.4991$/$0.4977$), so the choice does not drive any result.

Serve seed is irrelevant at $T=0$, as it must be. Two complete, independent greedy runs exist
at the \emph{same} seed. Comparing them isolates batching
non-determinism with everything else fixed: Gemma $I_{\text{within}}$ $0.9047$ vs.\ $0.9060$,
length-matched gap $+0.2110$ vs.\ $+0.2131$; Sarvam $0.8604$ vs.\ $0.8632$ and $+0.2200$ vs.\
$+0.2224$. Two jobs launched hours apart on different hardware agree to three decimal places.

Replicate agreement under greedy decoding is exact where it should be. Llama-4's two greedy replicates produce
\emph{identical} empty-trace counts (173/173 on the synthetic benchmark) and mean trace lengths
matching to two decimals on every benchmark.

The 300-task subset is representative. Ablation arms use 300 tasks per benchmark
rather than the full sweep. Invariance computed on the subset reproduces full-sweep invariance to
within $\pm 0.02$.

Benchmarks are resolved by content, not by name. Run directory names are length-capped and can
silently lose the benchmark suffix for long model identifiers, so all analyses identify a benchmark
by its \emph{language-count fingerprint} (21/12/15/16/11/23), which is unique across the suite and
survives any renaming. Main-sweep invariance recomputes to the reported values exactly
($\Delta = 0.0000$) under this resolver.

\section{Per-Cell Results}
\label{app:percell}

The pooled numbers in the main text average over cells of very different sizes, from XCOPA's
1{,}100 same-language pairs to FLORES-200's 63{,}000 cross-language ones. This appendix therefore
reports every cell separately, so that a reader can check whether any single benchmark or model
carries the result. It does not: all 48 length-matched gaps are positive, and at $T{=}0$ every
bootstrap interval on the raw gap excludes zero.

Figure~\ref{fig:percell} in the main text plots these cells; the numeric values, bootstrap
intervals and length-matched counterparts are tabulated below, under strict empty exclusion, a
symmetric cross-seed contrast and a 1{,}000-sample task-level bootstrap.

Every headline quantity in this paper is computed on
\emph{pooled pairs}, which weights a cell by how many language pairs it contributes.
Table~\ref{tab:permodel} instead prints $I_{\text{within}}$ and $I_{\text{cross}}$ macro-averaged
over the six benchmarks, so a reader who divides those two columns will not exactly recover the
printed $\tilde{I}$. We state both here so the difference is visible rather than discovered.
Recomputed from the printed model means, $\tilde{I}$ at $T{=}0$ is $0.746$ (Gemma), $0.715$
(Sarvam), $0.722$ (Qwen3) and $0.713$ (Llama-4), a span of $3.2$ points against the pooled $2.6$;
the ceiling correlation is $+0.51$ at $T{=}0.5$ and $+0.94$ at $T{=}0$ against the pooled $+0.43$
and $+0.97$; and the absolute-gap relative spread falls $87\% \rightarrow 44\%$ against the pooled
$81\% \rightarrow 34\%$. No conclusion in the paper depends on which aggregation is used: the
ordering, the direction and the collapse at $T{=}0$ are identical under both.

Dividing the two invariance columns of the $T{=}0$ block
below gives $\tilde{I}$ for each of the 24 cells, which is how \S\ref{sec:estimand} answers the
$n=4$ objection (Table~\ref{tab:percellret}). Because the division happens \emph{within} a cell,
trace length enters numerator and denominator identically and largely cancels, making this the
only benchmark-level comparison in the paper that length matching is not required to interpret.

\begin{table}[h]
\centering
\TableFontSize
\setlength{\tabcolsep}{4pt}
\begin{tabular}{@{}ll rr rrrr l@{}}
\toprule
\hdr Model & Benchmark & $n_{\text{w}}$ & $n_{\text{c}}$ & $I_{\text{within}}$ & $I_{\text{cross}}$ & Raw gap & Len-match & 95\% CI (raw gap) \\
\midrule
\multicolumn{9}{@{}l}{\emph{$T = 0.5$, matched replicates}} \\
\rw & FLORES-200 & 6{,}300 & 63{,}000 & 0.7327 & 0.6434 & 0.0894 & 0.0669 & [$+$0.0836, $+$0.0946] \\
\rw & XQuAD & 3{,}600 & 19{,}800 & 0.8626 & 0.7379 & 0.1248 & 0.0994 & [$+$0.1143, $+$0.1362] \\
\rw & XNLI & 4{,}500 & 31{,}500 & 0.8980 & 0.7439 & 0.1542 & 0.1450 & [$+$0.1476, $+$0.1600] \\
\rw & Belebele & 4{,}800 & 36{,}000 & 0.8141 & 0.6936 & 0.1205 & 0.1078 & [$+$0.1126, $+$0.1287] \\
\rw & XCOPA & 1{,}100 & 5{,}500 & 0.9018 & 0.7433 & 0.1585 & 0.1422 & [$+$0.1463, $+$0.1703] \\
\rw \multirow{-6}{*}{Gemma-3-27B} & Synthetic & 2{,}291 & 25{,}121 & 0.5731 & 0.4783 & 0.0948 & 0.0822 & [$+$0.0847, $+$0.1050] \\
\multirow{6}{*}{Sarvam-M}
& FLORES-200 & 6{,}237 & 62{,}261 & 0.5985 & 0.5669 & 0.0316 & 0.0266 & [$+$0.0271, $+$0.0360] \\
& XQuAD & 3{,}513 & 19{,}268 & 0.7364 & 0.6935 & 0.0428 & 0.0353 & [$+$0.0349, $+$0.0513] \\
& XNLI & 4{,}396 & 30{,}684 & 0.6079 & 0.5543 & 0.0535 & 0.0421 & [$+$0.0458, $+$0.0610] \\
& Belebele & 4{,}665 & 35{,}000 & 0.7727 & 0.7397 & 0.0330 & 0.0220 & [$+$0.0259, $+$0.0405] \\
& XCOPA & 1{,}082 & 5{,}393 & 0.6305 & 0.5604 & 0.0702 & 0.0617 & [$+$0.0585, $+$0.0833] \\
& Synthetic & 1{,}859 & 18{,}660 & 0.4362 & 0.4122 & 0.0239 & 0.0220 & [$+$0.0154, $+$0.0328] \\
\rw & FLORES-200 & 6{,}298 & 62{,}971 & 0.7028 & 0.5637 & 0.1390 & 0.1050 & [$+$0.1303, $+$0.1480] \\
\rw & XQuAD & 3{,}600 & 19{,}800 & 0.6051 & 0.4215 & 0.1836 & 0.1499 & [$+$0.1719, $+$0.1947] \\
\rw & XNLI & 4{,}500 & 31{,}500 & 0.6577 & 0.4885 & 0.1691 & 0.1354 & [$+$0.1589, $+$0.1797] \\
\rw & Belebele & 4{,}799 & 35{,}987 & 0.6372 & 0.5259 & 0.1112 & 0.0907 & [$+$0.1027, $+$0.1193] \\
\rw & XCOPA & 1{,}100 & 5{,}500 & 0.7131 & 0.6079 & 0.1052 & 0.0865 & [$+$0.0903, $+$0.1210] \\
\rw \multirow{-6}{*}{Qwen3-235B} & Synthetic & 1{,}822 & 16{,}978 & 0.5091 & 0.3784 & 0.1307 & 0.0933 & [$+$0.1126, $+$0.1493] \\
\multirow{6}{*}{Llama-4-Mav.}
& FLORES-200 & 6{,}300 & 63{,}000 & 0.7474 & 0.6458 & 0.1016 & 0.0879 & [$+$0.0967, $+$0.1066] \\
& XQuAD & 3{,}600 & 19{,}800 & 0.7978 & 0.6612 & 0.1366 & 0.1210 & [$+$0.1280, $+$0.1452] \\
& XNLI & 4{,}500 & 31{,}500 & 0.7835 & 0.6896 & 0.0939 & 0.0840 & [$+$0.0873, $+$0.1004] \\
& Belebele & 4{,}798 & 35{,}986 & 0.8070 & 0.7066 & 0.1003 & 0.0892 & [$+$0.0948, $+$0.1066] \\
& XCOPA & 1{,}100 & 5{,}500 & 0.7638 & 0.6580 & 0.1058 & 0.0964 & [$+$0.0912, $+$0.1203] \\
& Synthetic & 2{,}086 & 21{,}695 & 0.5910 & 0.5113 & 0.0797 & 0.0746 & [$+$0.0709, $+$0.0885] \\
\midrule
\multicolumn{9}{@{}l}{\emph{$T = 0$, greedy}} \\
\rw & FLORES-200 & 6{,}300 & 63{,}000 & 0.8749 & 0.6452 & 0.2297 & 0.1924 & [$+$0.2209, $+$0.2383] \\
\rw & XQuAD & 3{,}600 & 19{,}800 & 0.9632 & 0.7432 & 0.2200 & 0.2049 & [$+$0.2061, $+$0.2339] \\
\rw & XNLI & 4{,}500 & 31{,}500 & 0.9651 & 0.7440 & 0.2211 & 0.2150 & [$+$0.2138, $+$0.2284] \\
\rw & Belebele & 4{,}800 & 36{,}000 & 0.9520 & 0.6948 & 0.2572 & 0.2419 & [$+$0.2471, $+$0.2678] \\
\rw & XCOPA & 1{,}100 & 5{,}500 & 0.9745 & 0.7479 & 0.2266 & 0.2183 & [$+$0.2138, $+$0.2385] \\
\rw \multirow{-6}{*}{Gemma-3-27B} & Synthetic & 2{,}292 & 25{,}126 & 0.7062 & 0.4776 & 0.2287 & 0.2057 & [$+$0.2139, $+$0.2436] \\
\multirow{6}{*}{Sarvam-M}
& FLORES-200 & 6{,}267 & 62{,}436 & 0.8505 & 0.5741 & 0.2764 & 0.2355 & [$+$0.2705, $+$0.2821] \\
& XQuAD & 3{,}561 & 19{,}465 & 0.9399 & 0.7292 & 0.2107 & 0.1932 & [$+$0.1972, $+$0.2229] \\
& XNLI & 4{,}459 & 31{,}059 & 0.8883 & 0.5850 & 0.3033 & 0.2809 & [$+$0.2946, $+$0.3119] \\
& Belebele & 4{,}729 & 35{,}254 & 0.9625 & 0.7846 & 0.1779 & 0.1622 & [$+$0.1663, $+$0.1900] \\
& XCOPA & 1{,}097 & 5{,}471 & 0.8932 & 0.5779 & 0.3153 & 0.2994 & [$+$0.3033, $+$0.3280] \\
& Synthetic & 2{,}027 & 19{,}997 & 0.6446 & 0.4543 & 0.1903 & 0.1632 & [$+$0.1751, $+$0.2061] \\
\rw & FLORES-200 & 6{,}300 & 63{,}000 & 0.7400 & 0.5591 & 0.1809 & 0.1392 & [$+$0.1716, $+$0.1903] \\
\rw & XQuAD & 3{,}600 & 19{,}800 & 0.6661 & 0.4093 & 0.2568 & 0.2148 & [$+$0.2441, $+$0.2699] \\
\rw & XNLI & 4{,}499 & 31{,}490 & 0.6979 & 0.4815 & 0.2163 & 0.1750 & [$+$0.2063, $+$0.2267] \\
\rw & Belebele & 4{,}800 & 36{,}000 & 0.6683 & 0.5307 & 0.1376 & 0.1149 & [$+$0.1291, $+$0.1467] \\
\rw & XCOPA & 1{,}100 & 5{,}500 & 0.7488 & 0.5814 & 0.1674 & 0.1385 & [$+$0.1522, $+$0.1837] \\
\rw \multirow{-6}{*}{Qwen3-235B} & Synthetic & 1{,}820 & 16{,}921 & 0.5510 & 0.3789 & 0.1721 & 0.1280 & [$+$0.1514, $+$0.1923] \\
\multirow{6}{*}{Llama-4-Mav.}
& FLORES-200 & 6{,}300 & 63{,}000 & 0.9369 & 0.6394 & 0.2975 & 0.2800 & [$+$0.2910, $+$0.3036] \\
& XQuAD & 3{,}600 & 19{,}800 & 0.9204 & 0.6557 & 0.2647 & 0.2459 & [$+$0.2553, $+$0.2751] \\
& XNLI & 4{,}500 & 31{,}500 & 0.9417 & 0.6877 & 0.2540 & 0.2410 & [$+$0.2467, $+$0.2609] \\
& Belebele & 4{,}799 & 35{,}985 & 0.9273 & 0.7108 & 0.2165 & 0.1999 & [$+$0.2092, $+$0.2233] \\
& XCOPA & 1{,}100 & 5{,}500 & 0.9490 & 0.6557 & 0.2933 & 0.2826 & [$+$0.2777, $+$0.3086] \\
& Synthetic & 2{,}106 & 21{,}750 & 0.7277 & 0.5052 & 0.2226 & 0.2059 & [$+$0.2072, $+$0.2376] \\
\bottomrule
\end{tabular}
\caption{All 48 matched cells. Every gap is positive, and at $T{=}0$ every interval excludes zero.
\textbf{The intervals are task-level bootstraps on the raw gap, not on the length-matched value}, so
the length-matched column is not bracketed by them; since the raw gap exceeds the length-matched gap
in every cell and both are positive, the exclusion of zero carries to both.}
\end{table}

\section{Trace Length and Irreproducibility}
\label{app:predictors}

Qwen3-235B is a genuine outlier, retaining 32\% within-language dissimilarity even at $T{=}0$
against 9--14\% for the other three. On three models an attractive explanation is batch-dependent
expert routing in mixture-of-experts serving. \textbf{Llama-4 refutes it}: it is a 128-expert MoE
and the \emph{second most} reproducible model in the study ($1-I_{\text{within}} = 0.0995$),
statistically indistinguishable from dense Gemma ($0.0940$) and better than dense Sarvam
($0.1368$). We report the hypothesis alongside its refutation because on three models it looks
publishable and is wrong.

Testing candidate correlates \emph{per cell} across all 24 greedy cells, rather
than across four model means:

\begin{table}[h]
\centering
\TableFontSize
\setlength{\tabcolsep}{7pt}
\renewcommand{\arraystretch}{1.08}
\begin{tabular}{@{}lrr@{}}
\toprule
\hdr Predictor of $1 - I_{\text{within}}$ & $r$ & $R^2$ \\
\midrule
\rw Mean trace length & $+0.799$ & \keycell{\textbf{0.639}} \\
Mean agent iterations & $+0.665$ & 0.442 \\
\rw Collinearity (iterations $\leftrightarrow$ length) & $+0.425$ & --- \\
Mixture-of-experts indicator & --- & \warncell{no signal} \\
\bottomrule
\end{tabular}
\end{table}

\begin{table}[h]
\centering
\TableFontSize
\setlength{\tabcolsep}{6pt}
\renewcommand{\arraystretch}{1.08}
\begin{tabular}{@{}llccc@{}}
\toprule
\hdr Model & Arch. & Mean iter.\ (range) & Trace len.\ (range) & $1-I_{\text{within}}$ (range) \\
\midrule
\rw Gemma-3-27B & dense & 1.11--3.40 & 1.65--11.67 & 0.026--0.294 \\
Llama-4-Maverick & MoE & 1.00--2.19 & 2.37--7.64 & 0.051--0.272 \\
\rw Sarvam-M & dense & 1.03--2.99 & 1.41--10.92 & 0.038--0.355 \\
Qwen3-235B & MoE & 2.42--4.72 & 4.24--9.53 & \warncell{0.251--0.449} \\
\bottomrule
\end{tabular}
\caption{Per-cell predictors of greedy irreproducibility. Trace length is the strongest single
correlate and architecture carries no signal. The table also records a hypothesis the fourth model
rules out: on three models, Qwen3's outlier status is naturally attributed to batch-dependent expert routing in
MoE serving. Llama-4 refutes it: a 128-expert MoE that is the \emph{second most} reproducible
model in the study. At $n=4$ model means the ordering of length and iterations reverses, which is a
small-sample artifact and a reminder that model-level correlations on a handful of systems are
not evidence.}
\end{table}

\subsection{The three-level length manipulation}
\label{app:lengthcausal}

\begin{figure}[h]
\centering
\includegraphics[width=\textwidth]{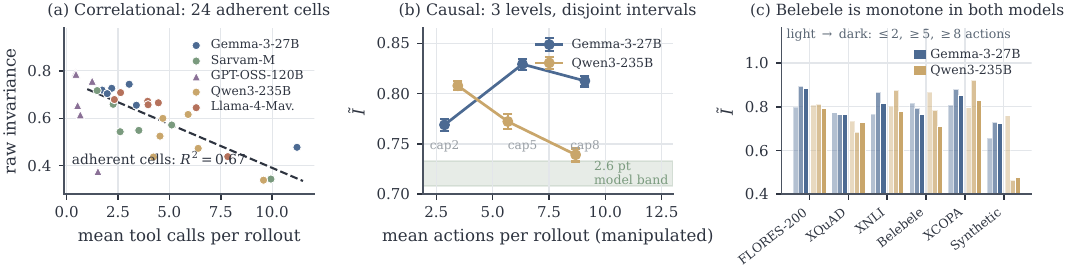}
\caption{Trace length, from correlation to cause. (a) The correlational picture across all 30
main-sweep cells, with the fit taken on the 24 protocol-adherent ones; triangles are GPT-OSS,
excluded. (b) The three-level manipulation, pooled, with the $2.6$-point across-model band shaded
for scale: $\tilde{I}$ moves further under length than across four frontier models. (c) Per
benchmark, the sign disagrees between models everywhere except Belebele.}
\label{fig:length}
\end{figure}

Everything above is correlational, and the correlation is strong enough
($R^2 = 0.67$ across the 24 adherent cells, $0.92$ within Gemma alone; Figure~\ref{fig:length}a)
that a difficulty account and a length account are observationally equivalent. We therefore
manipulated length directly. Three prompt-level arms (\emph{at most 2} actions, \emph{at least
5}, and \emph{at least 8}) were run at $T{=}0$ on two models with task, language, seed, decoding and
token budget held fixed, for $72$ cells and $271{,}200$ rollouts. Three levels rather than two
matters: a monotone move across three points is hard to attribute to an uncontrolled covariate,
and it also lets a non-monotone response be \emph{seen} rather than averaged away.

The manipulation took. Mean actions per rollout moves roughly threefold in both
models, Gemma $2.86 \rightarrow 6.32 \rightarrow 9.09$ and Qwen3
$3.45 \rightarrow 5.67 \rightarrow 8.68$, monotonically and in the instructed direction. Empty
traces are $0\%$ in every cell except Qwen3's synthetic ($21\%$, the pre-existing low-resource
adherence collapse of \S\ref{sec:correctness}), so the arms are not confounded with adherence.

\begin{table}[h]
\centering
\TableFontSize
\setlength{\tabcolsep}{6pt}
\renewcommand{\arraystretch}{1.06}
\begin{tabular}{@{}llrrrrl@{}}
\toprule
\hdr Model & Arm & Mean actions & $I_{\text{within}}$ & $I_{\text{cross}}$ & $\tilde{I}$ & 95\% CI \\
\midrule
\rw & $\leq 2$ actions & 2.86 & 0.9380 & 0.7212 & \keycell{0.7689} & [0.7632, 0.7751] \\
\rw & $\geq 5$ actions & 6.32 & 0.8817 & 0.7311 & 0.8292 & [0.8233, 0.8348] \\
\rw \multirow{-3}{*}{Gemma-3-27B} & $\geq 8$ actions & 9.09 & 0.8240 & 0.6696 & \keycell{0.8126} & [0.8070, 0.8179] \\
\multirow{3}{*}{Qwen3-235B}
& $\leq 2$ actions & 3.45 & 0.7940 & 0.6414 & \keycell{0.8078} & [0.8032, 0.8125] \\
& $\geq 5$ actions & 5.67 & 0.9813 & 0.7578 & 0.7722 & [0.7644, 0.7799] \\
& $\geq 8$ actions & 8.68 & 0.9653 & 0.7136 & \keycell{0.7392} & [0.7325, 0.7454] \\
\bottomrule
\end{tabular}
\caption{The three-level trace-length dose--response, at $T{=}0$ on 300 tasks with two replicates
per arm. The blue cells are the extreme arms, whose intervals are disjoint in both models: $\tilde{I}$ moves $6.0$ points in Gemma and $6.9$ in Qwen3 under a $3\times$ change
in length, against a $2.6$-point band across the four frontier models. Qwen3 is monotone
decreasing on all three points; Gemma is not, and its anomaly is the \emph{shortest} arm.}
\label{tab:lengthcausal}
\end{table}

Two things follow, and the second overturns a natural assumption. First, in Qwen3 the response is textbook, with longer traces, a larger raw gap and lower
retention, monotone on all three points, so the $R^2 = 0.67$ correlation is at least partly
causal rather than a difficulty proxy. Second, and against a natural intuition,
\textbf{the normalisation does not absorb length}. Because $\tilde{I}$ divides $I_{\text{cross}}$
by $I_{\text{within}}$ \emph{within} a cell, it is natural to argue that length enters numerator
and denominator together and cancels. It does not. The ratio moves $6.0$ and $6.9$ points with
disjoint intervals, which is \emph{larger} than the entire across-model band the paper's central
result rests on (Figure~\ref{fig:regime}c). Every benchmark-level statement that assumed
$\tilde{I}$ was length-neutral is withdrawn, and length is treated throughout as a first-order
driver rather than a nuisance parameter.

The sign is model-dependent, so no universal claim is made. Gemma is non-monotone in
both $\tilde{I}$ and the raw gap, and the anomaly sits at the \emph{short} end: constraining it to
at most two actions gives it the lowest retention of the three arms ($0.7689$ against $0.8292$ at
$\geq 5$). One reading, offered as a hypothesis rather than a result, is that at two actions a
model has no room to recover from a divergent opening move, whereas by five it has slack to
reconverge; that predicts a minimum at the short end, which is what Gemma shows and which Qwen3,
whose shortest arm already averages $3.45$ actions, may simply never reach. Two models and
three points cannot establish a U-shape. What is established is that the effect is
\textbf{model-dependent in sign}, which by itself defeats any universal statement about length.
Per benchmark (Figure~\ref{fig:length}c), Belebele is the one cell monotone decreasing in
\emph{both} models ($0.818 \rightarrow 0.765$ and $0.868 \rightarrow 0.710$) and is the cleanest
single instance of the causal effect.

Four caveats travel with this. The instruction is a \emph{soft} constraint, since ``at most 2'' and
``at least 5'' are prompt text rather than a decoding limit, so compliance is partial and the arms overlap in the
tails; the realised means in Table~\ref{tab:lengthcausal} are what the manipulation actually
achieved. Three levels is the minimum at which monotonicity is testable, and Gemma fails it, so a
four- or five-level ladder is needed to separate a U-shape from noise. Two models cannot settle
the sign. And every capped arm sits above its own uncapped published baseline, which must
\emph{not} be read as ``capping improves retention'': these arms use the 300-task subset while the
published baselines use the full sweep, and the across-benchmark spread within a single model
reaches 14 points. The arms are comparable \emph{to each other}, which is what the design requires.

\section{Uncorrected Metrics}
\label{app:raw}

These are the numbers a pipeline without the Appendix~\ref{app:confounds} corrections would
report. They are included so that the size of each correction is auditable, \textbf{not} as
results.

\begin{table}[h]
\centering
\TableFontSize
\setlength{\tabcolsep}{5pt}
\begin{tabular}{@{}lrrrrrr@{}}
\toprule
\hdr Model & FLORES & XQuAD & XNLI & Belebele & XCOPA & Synthetic \\
\midrule
\multicolumn{7}{@{}l}{\emph{Raw invariance (no baseline, no length match, no empty exclusion)}} \\
Gemma-3-27B & 0.6550 & 0.7189 & 0.7265 & 0.7040 & 0.7435 & 0.4777 \\
Sarvam-M & 0.5717 & 0.6590 & 0.5433 & 0.7168 & 0.5490 & 0.3437 \\
\rw GPT-OSS-120B & 0.3757 & 0.6152 & \textbf{0.7920} & 0.6549 & \textbf{0.7866} & 0.7572 \\
Qwen3-235B & 0.5997 & 0.4368 & 0.4732 & 0.5248 & 0.6162 & 0.3391 \\
Llama-4-Maverick & 0.6654 & 0.6791 & 0.6717 & 0.7083 & 0.6571 & 0.4383 \\
\midrule
\multicolumn{7}{@{}l}{\emph{Parse-failure rate}} \\
Gemma-3-27B & 0.0\% & 0.0\% & 0.0\% & 0.0\% & 0.0\% & 0.3\% \\
Sarvam-M & 0.7\% & 1.4\% & 1.5\% & 1.5\% & 0.6\% & 14.7\% \\
\rw GPT-OSS-120B & 51.1\% & 75.5\% & \textbf{88.6\%} & 79.9\% & \textbf{88.5\%} & 85.7\% \\
Qwen3-235B & 0.0\% & 0.0\% & 0.0\% & 0.0\% & 0.0\% & 20.7\% \\
Llama-4-Maverick & 0.0\% & 0.0\% & 0.0\% & 0.0\% & 0.0\% & 7.7\% \\
\midrule
\multicolumn{7}{@{}l}{\emph{Mean tool calls per rollout}} \\
Gemma-3-27B & 3.39 & 1.72 & 2.19 & 1.97 & 3.04 & 11.21 \\
Sarvam-M & 5.11 & 2.26 & 2.60 & 1.48 & 3.51 & 9.94 \\
\rw GPT-OSS-120B & 1.52 & 0.66 & 0.43 & 0.53 & 0.45 & 1.24 \\
Qwen3-235B & 4.68 & 4.22 & 6.40 & 4.54 & 5.91 & 9.58 \\
Llama-4-Maverick & 4.46 & 2.32 & 3.94 & 2.62 & 3.96 & 7.84 \\
\midrule
\multicolumn{7}{@{}l}{\emph{Median response length (characters)}} \\
Gemma-3-27B & 1{,}004 & 362 & 906 & 1{,}121 & 1{,}067 & 5{,}679 \\
Sarvam-M & 1{,}511 & 658 & 1{,}299 & 642 & 1{,}302 & 3{,}265 \\
\rw GPT-OSS-120B & \textbf{117} & \textbf{31} & \textbf{33} & \textbf{33} & \textbf{38} & 463 \\
Qwen3-235B & 971 & 1{,}092 & 1{,}375 & 1{,}742 & 1{,}178 & 4{,}653 \\
Llama-4-Maverick & 1{,}655 & 880 & 2{,}079 & 1{,}564 & 1{,}622 & 4{,}850 \\
\bottomrule
\end{tabular}
\caption{Bold entries mark the pathology of \S\ref{sec:correctness}: GPT-OSS's two
\emph{highest} raw invariance scores sit on its two highest parse-failure rates, and its median
response is one short sentence.}
\end{table}

A trace consisting of a single \texttt{Finish} call scores perfect
invariance while demonstrating no agentic behaviour. These are common: Sarvam produces them on
8{,}879 of 14{,}400 Belebele rollouts (61.7\%), Gemma on 7{,}215 of 14{,}268 XQuAD rollouts.
Any invariance number that does not exclude them is partly a measure of degeneracy.

Tool-name hallucination is negligible ($\leq 0.11\%$ of calls for
every model). Across the main sweep, 1{,}412{,}535 tool calls were emitted, of which 389 were
non-canonical (most commonly \texttt{calculate} for \texttt{Calc}). The action alphabet is
therefore genuinely shared, which is what makes cross-lingual trace comparison well-posed.

\section{GPT-OSS-120B Failure Taxonomy}
\label{app:gptoss}

This is the paper's clearest single demonstration that an evaluation artifact can be mistaken for
a model property, so we document it in full. The question the table below answers is narrow but
consequential: when a model's measured accuracy moves by more than an order of magnitude under an
intervention that touches neither its weights nor its decoding, which of the two numbers was ever
measuring capability?

\begin{table}[h]
\centering
\TableFontSize
\setlength{\tabcolsep}{5pt}
\renewcommand{\arraystretch}{1.08}
\begin{tabular}{@{}lrrrrrr@{}}
\toprule
\hdr & \multicolumn{3}{c}{Adherence (GPT-OSS-120B)} & \multicolumn{3}{c}{Correctness (gold benchmarks)} \\
\cmidrule(lr){2-4} \cmidrule(lr){5-7}
\hdr Arm & Empty & Parse fail & Usable pairs & Scorable & Accuracy & Acc.\,$|$\,scorable \\
\midrule
\rw 0-shot (baseline) & \warncell{74.4\%} & 59.5\% & \warncell{10.4\%} & 2.1\% & \warncell{0.0174} & \textbf{0.8127} \\
2-shot & 28.7\% & 21.7\% & 53.3\% & 58.4\% & 0.4421 & 0.7574 \\
\rw 4-shot & \goodcell{\textbf{27.8\%}} & \textbf{19.7\%} & \goodcell{\textbf{55.6\%}} & \textbf{61.2\%} & \goodcell{\textbf{0.4539}} & \textbf{0.7418} \\
\bottomrule
\end{tabular}
\caption{Format exemplars repair measurement, not capability. Measured accuracy rises
$26\times$ (orange to green) while accuracy among scorable rollouts \emph{falls} slightly: the
0-shot $0.8127$ rests on $n=299$ heavily selected rollouts, the 4-shot $0.7418$ on $n=8{,}567$.
The effect saturates at two exemplars, which is itself informative: if the exemplars were teaching
the task rather than the format, four would beat two.}
\label{tab:fewshot}
\end{table}

Per-benchmark few-shot accuracy for GPT-OSS runs Belebele $0.023 \to 0.399 \to 0.362$;
XCOPA $0.026 \to 0.530 \to 0.579$; XNLI $0.017 \to 0.513 \to 0.570$; XQuAD
$0.007 \to 0.384 \to 0.394$ (0-shot $\to$ 2-shot $\to$ 4-shot). Qwen3, already legible, gains far
less: Belebele $0.782 \to 0.885 \to 0.892$, XCOPA $0.855 \to 0.919 \to 0.929$, XNLI
$0.611 \to 0.756 \to 0.750$.

\begin{table}[h]
\centering
\TableFontSize
\setlength{\tabcolsep}{5pt}
\begin{tabular}{@{}lrrrrr@{}}
\toprule
\hdr Benchmark & Rollouts & No trace & Non-empty & Names a tool & \% of non-empty \\
\midrule
\rw FLORES-200 & 20{,}937 & 10{,}692 & 9{,}842 & 9{,}745 & 99.0\% \\
XQuAD & 14{,}268 & 10{,}767 & 7{,}299 & 6{,}673 & 91.4\% \\
\rw XNLI & 37{,}350 & 33{,}095 & 29{,}700 & 26{,}672 & 89.8\% \\
Belebele & 14{,}400 & 11{,}500 & 8{,}614 & 7{,}248 & 84.1\% \\
\rw XCOPA & 1{,}100 & 974 & 804 & 772 & 96.0\% \\
Synthetic & 2{,}300 & 1{,}970 & 1{,}299 & 948 & 73.0\% \\
\midrule
\rw \textbf{Total} & \textbf{90{,}355} & \textbf{68{,}998} & \textbf{57{,}558} & \textbf{52{,}058} & \textbf{90.4\%} \\
\bottomrule
\end{tabular}
\caption{The final column is the share of \emph{non-empty} failures, and must never be quoted
without that qualifier. Over \emph{all} failures the recovery ceiling is
$52{,}058/68{,}998 = 75.4\%$; the residual 16.6\% are genuinely empty responses. Post-recovery
adherence would be 81.3\%, up from 23.6\%.}
\end{table}

Under strict empty exclusion, 89.93\% of GPT-OSS's within-language pairs vanish, leaving
$n_{\text{within}} = 24$ on XCOPA and $68$ on XNLI, and 3 of its 6 cells have intervals crossing
zero. Its apparent gap of $+0.0450$ measures the minority of cases where it followed the protocol
at all, which is why it is excluded from the headline pooling. \S\ref{sec:correctness} shows this
is a \emph{measurement} decision, not a capability judgement.

\section{Temperature Ladder}
\label{app:ladder}

The ladder exists because two points do not distinguish ``the gap grows at $T{=}0$'' from ``the
ceiling moves at $T{=}0$''. Five rungs on Gemma-3-27B and four on Sarvam-M, with model, task set
and seed fixed, separate the two: $I_{\text{cross}}$ is read \emph{within} a model across
temperature, which is the only comparison the single-replicate rungs support. Ceiling-corrected
columns exist only where two replicates were generated.

\begin{table}[h]
\centering
\TableFontSize
\setlength{\tabcolsep}{5pt}
\begin{tabular}{@{}lrrrrrrrr@{}}
\toprule
\hdr & \multicolumn{3}{c}{$I_{\text{cross}}$} & \multicolumn{2}{c}{Behaviour} & \multicolumn{3}{c}{Ceiling-corrected} \\
\cmidrule(lr){2-4} \cmidrule(lr){5-6} \cmidrule(lr){7-9}
$T$ & Raw & A: $\to$0.5 & B: 0.5$\to$ & Len. & Empty & $I_{\text{within}}$ & Gap & $\tilde{I}$ \\
\midrule
\multicolumn{9}{@{}l}{\emph{Gemma-3-27B}} \\
\rw 0.0 & 0.6635 & 0.6611 & 0.6632 & 3.48 & 0.04\% & 0.9110 & $+$0.2482 & 0.7276 \\
0.3 & 0.6632 & 0.6608 & 0.6631 & 3.45 & 0.04\% & --- & --- & --- \\
\rw 0.5 & 0.6608 & 0.6608 & 0.6608 & 3.44 & 0.04\% & 0.7957 & $+$0.1344 & 0.8311 \\
0.7 & 0.6565 & 0.6576 & 0.6597 & 3.39 & 0.04\% & 0.7623 & $+$0.1191 & 0.8438 \\
\rw 1.0 & 0.6546 & 0.6576 & 0.6576 & 3.38 & 0.05\% & --- & --- & --- \\
\midrule
\multicolumn{9}{@{}l}{\emph{Sarvam-M}} \\
0.0 & 0.6226 & 0.6162 & 0.6025 & 3.88 & 1.79\% & 0.8797 & $+$0.2572 & 0.7076 \\
\rw 0.3 & 0.6153 & 0.6099 & 0.6013 & 3.90 & 1.93\% & --- & --- & --- \\
0.5 & 0.5958 & 0.5958 & 0.5958 & 3.84 & 2.42\% & 0.6477 & $+$0.0506 & 0.9219 \\
\rw 0.7 & 0.5697 & 0.5772 & 0.5884 & 3.77 & 4.26\% & 0.5977 & $+$0.0400 & 0.9330 \\
\bottomrule
\end{tabular}
\caption{$T=0.3$ and $T=1.0$ are single-replicate by design, so no ceiling correction exists
there; the flatness claim rests on $I_{\text{cross}}$, used only as a within-model comparison
across temperature with model and task set fixed. Sarvam's $T=1.0$ arm was not generated.
\textbf{The $\tilde{I}$ column is computed on length-matched $I_{\text{cross}}$ whereas the
``Raw'' column is not}, so dividing the two neighbouring columns does not reproduce it: at
$T{=}0.7$ Gemma's raw ratio is $0.861$ against the length-matched $0.8438$ printed here.}
\end{table}

\section{English-Pivot Ablation}
\label{app:pivot}

The ablation ran in two campaigns under one protocol. Gemma-3-27B and Qwen3-235B were run first
(24 cells, 90{,}400 rollouts) and produced a \emph{post-hoc} head-room account of why the mandate
helped one and not the other. Llama-4-Maverick and Sarvam-M were then selected \emph{on that
account}, from their measured baseline pivot rate rather than from any outcome, and both
predictions were written into the analysis script before the compute was spent (36 cells,
135{,}600 rollouts, including an in-job baseline arm so that all three arms are matched by
construction rather than borrowed). \textbf{Sixty cells and 226{,}000 rollouts in total}, all
matched on seed 101, $T{=}0.5$, 4096 tokens and 300 tasks, with the same two intervention prompt
files used verbatim for all four models so the manipulation is textually identical rather than
merely equivalent.

\subsection{The pre-registered test, consolidated}

Table~\ref{tab:pivot-four} is the four-model result in one place. The upper two rows of each
block are the original campaign, the lower two the pre-registered extension; reading the
``Bench'' columns downward is the test.

\begin{table}[h]
\centering
\TableFontSize
\setlength{\tabcolsep}{4pt}
\renewcommand{\arraystretch}{1.08}
\begin{tabular}{@{}lrr rrc rrc@{}}
\toprule
\hdr & & & \multicolumn{3}{c}{\texttt{Translate} removed} & \multicolumn{3}{c}{\texttt{Translate} mandated} \\
\cmidrule(lr){4-6} \cmidrule(lr){7-9}
\hdr Model & Head-room & Moved & A & B & Bench & A & B & Bench \\
\midrule
\rw Sarvam-M$^\ddagger$ & \textbf{57.2} & $+38.6$ & $+0.0015$ & $+0.0064$ & 1/6 & $+0.0110$ & $+0.0190$ & \goodcell{\textbf{5/6}} \\
Gemma-3-27B & 30.4 & $+30.0$ & $\mathbf{-0.0432}$ & $\mathbf{-0.0348}$ & 4/6 & $\mathbf{+0.0635}$ & $\mathbf{+0.0674}$ & \goodcell{\textbf{5/6}} \\
\rw Llama-4-Maverick & 18.7 & $+17.4$ & $-0.0511$ & \warncell{$+0.0124$} & 4/6 & $+0.0242$ & $+0.0154$ & 3/6 \\
Qwen3-235B & 15.7 & $+14.7$ & $-0.0229$ & \warncell{$+0.0659$} & 4/6 & $-0.0105$ & $-0.0117$ & \warncell{2/6} \\
\bottomrule
\end{tabular}
\caption{All four ablated models, ordered by head-room; the two lower rows in each block were
predicted before generation. Head-room is $100$ minus the baseline first-action
\texttt{Translate} rate; ``Moved'' is the increase in that rate under the mandate. A and B are the
two length-matching directions on $I_{\text{cross}}$; orange marks a pooled estimate that fails the
both-directions test, in which case the per-benchmark counts are the honest read.
$^\ddagger$Sarvam's mandate reached only $81.4\%$ first-action compliance against $98.7$--$99.6\%$
elsewhere, so part of its small magnitude may be weak manipulation rather than a model property;
this is the largest single threat to the magnitude reading below.}
\label{tab:pivot-four}
\end{table}

Direction is predicted; magnitude is not. Ranked by benchmarks where the mandate helps
with both directions agreeing, the four models give $5/6$, $5/6$, $3/6$ and $2/6$, non-increasing
in head-room, on four independent systems, with the two new points fixed in advance. The pooled
magnitudes do not follow: Gemma $+0.0655$, Llama-4 $+0.0198$, Sarvam $+0.0150$, Qwen3 $-0.0111$.
Sarvam has the most head-room and \emph{moved furthest} ($+38.6$ points against Gemma's $+30.0$),
yet gained $4.4\times$ less, and Llama-4 is third on head-room but second on magnitude. We
therefore report head-room as governing \emph{whether} the mandate helps, not how much, and offer
no prescription.

Llama-4's removal arm is the third instance of the raw-versus-matched inversion, and is the cleanest
case in the paper. Raw $I_{\text{cross}}$ \emph{rises} $0.6484 \rightarrow 0.7038$ ($+0.0554$),
which read alone would reverse \S\ref{sec:mechanism} outright; removal shortens its traces by
$24\%$ ($4.08 \rightarrow 3.09$ actions), and matching to the baseline length distribution flips
the sign to $-0.0511$. That the inversion recurs on a third model, in a campaign designed after
the first two, is why every comparison in this paper is length-matched in both directions.

Two pooled estimates fail the both-directions test, for a reason worth stating. Direction B reweights the
baseline onto the arm's length distribution, which is unreliable when the arm populates a bin the
baseline barely reaches. Llama-4's removal arm puts $1{,}593$ cross-language pairs in the
shortest bin where its baseline has only $582$, and its mandate arm only $65$; Sarvam, by
contrast, has at least $10{,}378$ pairs in every bin of every arm. Sarvam's estimates are
therefore the best-supported in the set, which matters because Sarvam is the model that supplies
the negative result below.

\subsection{The removal finding does not generalise, and that is the informative part}

Across two models, removal lowered length-matched cross-lingual agreement in both. Across four it
does not. \textbf{Llama-4 replicates it per benchmark}, 4/6 negative with both directions
agreeing, even though its pooled estimate fails the both-directions test.
\textbf{Sarvam-M contradicts it}: only 1/6 benchmarks is negative and the pooled estimate is
$+0.0015$ / $+0.0064$, which at one seed per arm should be read as \emph{no effect} rather than as
a small positive one. The coherent reading is that the pivot is load-bearing \emph{in proportion to
how much a model uses it}: Sarvam opens with \texttt{Translate} on $42.8\%$ of non-English
rollouts, roughly half of the other three, so there is less pivoting to remove and removal costs it
nothing. That is a dose--response refinement of the mechanism rather than a refutation of it, but a
flat claim that removal lowers agreement ``in models'' is not available, and we do not make one.

\subsection{Per-benchmark length-matched deltas}

Each entry is the pair of length-matched deltas (direction A / direction B) for that benchmark, and
a cell counts toward the ``Net'' row only when the two directions agree in sign.

\begin{table}[h]
\centering
\TableFontSize
\setlength{\tabcolsep}{4.5pt}
\begin{tabular}{@{}lrrrr@{}}
\toprule
\hdr Benchmark & Gemma removed & Gemma mandated & Qwen3 removed & Qwen3 mandated \\
\midrule
\rw FLORES-200 & $-0.054$ / $-0.054$ & $+0.037$ / $+0.076$ & $-0.006$ / $+0.262$ $\times$ & $-0.010$ / $-0.001$ \\
XQuAD & $+0.009$ / $+0.007$ & $+0.133$ / $+0.215$ & $-0.027$ / $-0.014$ & $+0.088$ / $+0.151$ \\
\rw XNLI & $-0.245$ / $-0.085$ & $+0.021$ / $+0.022$ & $-0.170$ / $-0.038$ & $-0.062$ / $-0.043$ \\
Belebele & $-0.047$ / $-0.031$ & $+0.158$ / $+0.167$ & $-0.122$ / $-0.072$ & $+0.046$ / $+0.065$ \\
\rw XCOPA & $-0.227$ / $-0.168$ & $+0.016$ / $+0.014$ & $-0.308$ / $-0.210$ & $-0.166$ / $-0.041$ \\
Synthetic$^\dagger$ & $+0.051$ / $+0.052$ & $-0.022$ / $-0.017$ & $+0.089$ / $+0.114$ & $-0.149$ / $-0.062$ \\
\midrule
\rw \textbf{Net} & \textbf{4/6 negative} & \textbf{5/6 positive} & \textbf{4/6 negative} & 2/6 positive \\
\midrule
\hdr Benchmark & Llama-4 removed & Llama-4 mandated & Sarvam removed & Sarvam mandated \\
\midrule
\rw FLORES-200 & $-0.133$ / $-0.069$ & $+0.034$ / $+0.037$ & $-0.067$ / $-0.050$ & $+0.039$ / $+0.048$ \\
XQuAD & $+0.065$ / $+0.079$ & $+0.031$ / $+0.098$ & $+0.002$ / $+0.001$ & $+0.009$ / $+0.081$ \\
\rw XNLI & $-0.104$ / $-0.060$ & $-0.059$ / $-0.007$ & $+0.052$ / $+0.066$ & $+0.068$ / $+0.074$ \\
Belebele & $-0.024$ / $+0.085$ $\times$ & $+0.016$ / $+0.052$ & $+0.005$ / $+0.005$ & $+0.001$ / $+0.015$ \\
\rw XCOPA & $-0.088$ / $-0.051$ & $-0.016$ / $+0.018$ $\times$ & $+0.091$ / $+0.112$ & $+0.093$ / $+0.105$ \\
Synthetic$^\dagger$ & $-0.006$ / $-0.004$ & $-0.020$ / $-0.017$ & $+0.017$ / $+0.017$ & $-0.045$ / $-0.038$ \\
\midrule
\rw \textbf{Net} & \textbf{4/6 negative} & 3/6 positive & 1/6 negative & \textbf{5/6 positive} \\
\bottomrule
\end{tabular}
\caption{$\times$ marks a cell where the two matching directions disagree in sign and which
therefore does not count. Sarvam's per-benchmark agreement is $12/12$, covering every cell, both
arms and both directions, against Llama-4's $10/12$. Two patterns cut across models. \textbf{Synthetic
inverts under the mandate in all four}, and it is also one of the two benchmarks where removal
\emph{raises} agreement in three of them; both are consistent with its far lower chance floor
($0.351$ against $0.57$--$0.64$; Appendix~\ref{app:chancefloor}) meaning it measures something
different from the natural benchmarks. \textbf{XQuAD is the other dissenter for removal},
positive in three of four models. On every other benchmark, removal is negative wherever a model
pivots heavily. $^\dagger$\textbf{The Qwen3 synthetic cell additionally carries a truncation confound
and should be discounted}: mean trace length is 9.53 (baseline), 10.44 (removed) and 3.69
(mandated), with truncation 21.6\%, 34.8\% and 6.6\%. The 20.7--21.1\% empty rate there is the
pre-existing low-resource adherence collapse and is stable across arms.}
\end{table}

\subsection{Tool redistribution under the ablation}

Table~\ref{tab:toolshare} gives each arm's tool distribution. Calls to a tool that was not offered
number 6 in $\sim$500{,}000 ($0.01\%$), so the alphabet restriction was respected in every model.

\begin{table}[h]
\centering
\TableFontSize
\setlength{\tabcolsep}{7pt}
\renewcommand{\arraystretch}{1.08}
\begin{tabular}{@{}ll rrrrr@{}}
\toprule
\hdr Model & Arm & \texttt{Search} & \texttt{Calc} & \texttt{Translate} & \texttt{Summarize} & \texttt{Finish} \\
\midrule
\rw & baseline & 10.2 & 21.3 & 28.3 & 11.9 & 28.2 \\
\rw & \texttt{Translate} removed & \goodcell{25.7} & \goodcell{25.7} & \keycell{0.0} & 15.9 & 32.7 \\
\rw \multirow{-3}{*}{Gemma-3-27B} & \texttt{Translate} mandated & 9.5 & 15.0 & 30.2 & 18.6 & 26.7 \\
\midrule
& baseline & 5.0 & 10.7 & 50.5 & 20.2 & 13.6 \\
& \texttt{Translate} removed & 15.8 & 15.9 & \keycell{0.0} & \goodcell{\textbf{54.4}} & 13.9 \\
\multirow{-3}{*}{Qwen3-235B} & \texttt{Translate} mandated & 6.3 & 4.6 & 50.1 & 20.3 & 18.7 \\
\midrule
\rw & baseline & 10.2 & 11.3 & 35.6 & 18.8 & 24.1 \\
\rw & \texttt{Translate} removed & 14.8 & 13.8 & \keycell{0.0} & \goodcell{\textbf{39.8}} & 31.7 \\
\rw \multirow{-3}{*}{Llama-4-Maverick} & \texttt{Translate} mandated & 14.2 & 10.7 & 36.9 & 17.5 & 20.6 \\
\midrule
& baseline & 23.5 & 16.1 & 18.0 & 18.0 & 24.2 \\
& \texttt{Translate} removed & \goodcell{\textbf{32.2}} & 20.2 & \keycell{0.0} & 21.7 & 25.9 \\
\multirow{-3}{*}{Sarvam-M} & \texttt{Translate} mandated & 22.0 & 13.5 & 25.4 & 16.5 & 22.3 \\
\bottomrule
\end{tabular}
\caption{Share of all tool calls (\%) by model and arm. Blue marks the withdrawn tool, green where
its budget went. Removal is absorbed by \texttt{Summarize} in Qwen3 and Llama-4, by
\texttt{Search} in Sarvam, and split between \texttt{Search} and \texttt{Calc} in Gemma.}
\label{tab:toolshare}
\end{table}

The freed budget is reallocated \emph{differently}, and the difference tracks the invariance
result. Qwen3 sends $68\%$ of the withdrawn share to \texttt{Summarize}
($20.2 \rightarrow 54.4$) and Llama-4 $59\%$ ($18.8 \rightarrow 39.8$), so both substitute another
English-producing operation. Gemma splits, $55\%$ to \texttt{Search} and $16\%$ to \texttt{Calc}.
Sarvam sends only $20\%$ to \texttt{Summarize} and $48\%$ to \texttt{Search}
($23.5 \rightarrow 32.2$). In every model the gains reconcile to the withdrawn share to within
$0.2$ points, so this is a genuine reallocation rather than a counting artifact. The two models
that lean hardest on the pivot replace it with the one remaining tool that also emits English; the
model where removal costs nothing reaches instead for retrieval.

In the reasoning-language control, \texttt{Thought:} text in the task's script, by arm:
Gemma 0.09\% (unconstrained), 0.06\% (think-in-English), 0.79\% (think-in-task-language); Sarvam
0.16\% / 0.04\% / 0.08\%. Mean ASCII fraction of \texttt{Thought:} text on non-Latin-script arms
is 0.986--0.993 in every condition. The think-in-English arm's length-matched effect is $+0.011$
(Gemma) and $+0.014$ (Sarvam), with both matching directions agreeing.

\section{Self-Consistency Voting}
\label{app:voting}

Voting is the most obvious repair a practitioner would reach for, so it deserves a careful report
rather than silence. Because majority voting reduces variance, it must raise same-language
agreement as well as cross-language agreement, and C4 says a gain read without its ceiling is
uninterpretable. Testing $\tilde{I}$ under voting needs $2k$ replicates so that \emph{both} sides
of the ratio are themselves voted. We therefore ran ten replicates, and the question is now
settled.

Voting is a variance reducer, not a retention improver. With vote A (seeds 701--705) and vote B (seeds 706--710) at $T{=}0.7$ on 300 tasks, $\tilde{I}$ at
$k{=}5$ is formable for the first time. For Gemma-3-27B it falls from $0.8481$ $[0.8438, 0.8527]$
at $k{=}1$ to $0.8323$ $[0.8274, 0.8372]$ at $k{=}5$; for Sarvam-M from $0.9296$
$[0.9238, 0.9353]$ to $0.9109$ $[0.9062, 0.9159]$. Both intervals are disjoint. The mechanism is
visible in the components: voting raises $I_{\text{within}}$ by $+0.078$ and $+0.071$ against
$I_{\text{cross}}$ by $+0.052$ and $+0.053$: a five-way vote makes a model agree with
\emph{itself} more than it makes it agree \emph{across languages}. Two qualifications travel with
this. The effect is \textbf{small} ($-0.016$, $-0.019$), and the intervals are disjoint partly
because task-level bootstraps within a model are narrow by construction. And it is measured at
$T{=}0.7$ only, since voting is undefined at $T{=}0$ where all replicates coincide, so it
cannot be checked at the temperature where the retention regularity is defined. Both readings are
true: raw cross-language agreement does rise $+0.053$, and a practitioner whose objective is raw
agreement rather than retention relative to a model's own ceiling will obtain it.

The table below reports the \emph{five}-replicate run (seeds 701--705, $T{=}0.7$, majority vote over
tool sequences), which measures raw $I_{\text{cross}}$ under voting. The ceiling-corrected result
above uses the separate \emph{ten}-replicate run, since forming $\tilde{I}$ needs both sides of the
ratio voted; the two are different experiments and their numbers should not be compared directly.

\begin{table}[h]
\centering
\TableFontSize
\setlength{\tabcolsep}{6pt}
\begin{tabular}{@{}llrrrrr@{}}
\toprule
\hdr Model & Benchmark & $k{=}1$ & $k{=}3$ & $k{=}5$ & $\Delta(k_5{-}k_1)$ & Unanimous \\
\midrule
\multirow{7}{*}{Gemma-3-27B}
& FLORES-200 & 0.6319 & 0.6718 & 0.7283 & $+$0.0964 & 13.8\% \\
& XQuAD & 0.7276 & 0.7502 & 0.7703 & $+$0.0427 & 44.5\% \\
& XNLI & 0.7411 & 0.7522 & 0.7585 & $+$0.0174 & 47.3\% \\
& Belebele & 0.6885 & 0.7134 & 0.7298 & $+$0.0414 & 27.2\% \\
& XCOPA & 0.7416 & 0.7468 & 0.7515 & $+$0.0099 & 46.9\% \\
& Synthetic & 0.4688 & 0.4693 & 0.4738 & $+$0.0050 & 2.0\% \\
& \textbf{pooled} & \textbf{0.6428} & \textbf{0.6645} & \textbf{0.6898} & $\mathbf{+0.0470}$ & \textbf{27.8\%} \\
\midrule
\multirow{7}{*}{Sarvam-M}
& Belebele & 0.6914 & 0.7423 & 0.7957 & $+$0.1043 & 16.9\% \\
& XQuAD & 0.6534 & 0.6890 & 0.7418 & $+$0.0884 & 9.2\% \\
& FLORES-200 & 0.5361 & 0.5463 & 0.5862 & $+$0.0501 & 0.4\% \\
& XNLI & 0.5347 & 0.5434 & 0.5714 & $+$0.0367 & 1.3\% \\
& XCOPA & 0.5364 & 0.5369 & 0.5651 & $+$0.0288 & 0.5\% \\
& Synthetic & 0.3891 & 0.3846 & 0.3801 & $-$0.0090 & 0.1\% \\
& \textbf{pooled} & \textbf{0.5568} & \textbf{0.5706} & \textbf{0.6040} & $\mathbf{+0.0472}$ & \textbf{5.1\%} \\
\bottomrule
\end{tabular}
\caption{The gain is largest at \emph{intermediate} replicate unanimity, not at either extreme:
mean $\Delta$ is $+0.096$ in the 5--20\% unanimity band against $+0.022$ below 5\% and $+0.028$
above 20\%. Where replicates already agree there is nothing to vote away, and where they disagree
completely there is no majority to find, so the relationship is an inverted U rather than a
monotone one; the pooled linear correlation is correspondingly near zero ($r = -0.08$). Sarvam
never exceeds 16.9\% unanimity and therefore sits entirely on the rising limb. \textbf{Pooled rows
are computed over pairs, not as the mean of the cells above them}, so they need not equal that
mean (Gemma: $+0.0470$ pooled against $+0.0355$ macro-averaged), the same distinction noted for
Table~\ref{tab:permodel} in Appendix~\ref{app:percell}.
Because voting also raises $I_{\text{within}}$ and $k=5$ cannot supply two independent votes,
these gains are \textbf{not} evidence that voting improves cross-lingual agreement
(\S\ref{sec:correctness}).}
\end{table}

\FloatBarrier
\section{Correctness}
\label{app:correctness}

Correctness enters this paper for one reason: to test whether trace divergence is merely
cosmetic. It is not, and the detail below also shows why invariance cannot be substituted for
accuracy as a cheap evaluation signal.

Gold is available for XQuAD (extractive, per-language), XNLI (label), Belebele (MCQ) and XCOPA
(MCQ). Scoring is deliberately lenient (first standalone digit for MCQ, first NLI label
mentioned, normalised containment for extractive QA), which biases \emph{against} finding a
spurious accuracy--invariance relationship.

\begin{table}[h]
\centering
\TableFontSize
\setlength{\tabcolsep}{6pt}
\begin{tabular}{@{}llrrrr@{}}
\toprule
\hdr Model & Benchmark & Mean & Min & Max & Spread \\
\midrule
\multirow{4}{*}{Llama-4-Maverick} & XCOPA & 0.949 & 0.840 & 1.000 & 0.160 \\
& Belebele & 0.882 & 0.848 & 0.933 & 0.086 \\
& XNLI & 0.718 & 0.659 & 0.784 & 0.125 \\
& XQuAD & 0.695 & 0.458 & 0.822 & 0.364 \\
\multirow{4}{*}{Gemma-3-27B} & XCOPA & 0.901 & 0.540 & 0.980 & \textbf{0.440} \\
& Belebele & 0.802 & 0.726 & 0.837 & 0.111 \\
& XNLI & 0.699 & 0.622 & 0.780 & 0.157 \\
& XQuAD & 0.615 & 0.462 & 0.860 & 0.399 \\
\multirow{4}{*}{Qwen3-235B} & XCOPA & 0.853 & 0.330 & 0.990 & \textbf{0.660} \\
& Belebele & 0.769 & 0.653 & 0.917 & 0.263 \\
& XNLI & 0.606 & 0.402 & 0.790 & 0.388 \\
& XQuAD & 0.501 & 0.352 & 0.638 & 0.287 \\
\multirow{4}{*}{Sarvam-M} & Belebele & 0.807 & 0.452 & 0.943 & \textbf{0.491} \\
& XCOPA & 0.785 & 0.450 & 0.940 & 0.490 \\
& XNLI & 0.648 & 0.503 & 0.767 & 0.263 \\
& XQuAD & 0.485 & 0.278 & 0.789 & \textbf{0.511} \\
\bottomrule
\end{tabular}
\caption{Per-language accuracy range within each model$\times$benchmark cell. GPT-OSS is omitted
here: its accuracy is adherence-limited (\S\ref{sec:correctness}) and reported separately.}
\end{table}

The accuracy--invariance relation reverses in five cells: Gemma$\cdot$XNLI $-0.477$,
Llama-4$\cdot$Belebele $-0.669$, Llama-4$\cdot$XNLI $-0.376$, Qwen3$\cdot$Belebele $-0.299$,
Qwen3$\cdot$XNLI $-0.129$. Median across the 20 adherent cells is $+0.513$.

\section{The Chance Floor}
\label{app:chancefloor}

$S$ is a normalised matching-block similarity over a five-symbol alphabet, so two traces that
answer \emph{different} tasks still agree substantially by construction. Under the standard
attenuation model $S_{\text{obs}} = c + (1-c)\,S_{\text{true}}$, the difference
$\Delta_{\text{obs}} = (1-c)\,\Delta_{\text{true}}$ rescales cleanly and its ordering is invariant
to $c$, but the ratio $\tilde{I}_{\text{obs}} = (c + (1-c)I_{\text{cross}})/(c + (1-c)I_{\text{within}})$
is biased toward $1$ and differentially so across models. The estimand we argue for is therefore
strictly more exposed to an uncorrected floor than the quantity it replaces.

Within each (model, benchmark, condition) cell we permute the task\,$\rightarrow$\,trace
assignment \emph{within} a language arm and recompute the ordinary cross-language $S$. This is the
exact null ``what does $S$ score when two traces answer different tasks?'', and it preserves the
language, the model, the length distribution and the empty-trace pattern of the arm while
destroying only the task correspondence. Everything else is held identical to the main estimator:
the same $S$, the same strict empty exclusion, the same cross-seed pairing. Because $c$ is
length-dependent it is estimated \emph{within} length bin and reweighted to the observed
cross-language bin distribution, exactly as the length-matched estimator does. We use 200
permutations and a fixed seed, giving 1.07M--12.6M null pairs per cell over \textbf{66 cells}.

The floor is high, and benchmark-specific. Mean $c = 0.561$ at $T{=}0$ (per-cell range $0.266$--$0.753$) and $0.559$ at $T{=}0.5$. By
benchmark at $T{=}0$: Belebele $0.644$, XCOPA $0.626$, XNLI $0.591$, FLORES-200 $0.585$, XQuAD
$0.570$, synthetic $0.351$: the ordering tracks trace length, since the synthetic benchmark has
by far the longest traces and correspondingly the lowest collision rate. Across the three P6-A2
models the range widens to $0.314$--$0.947$, and \textbf{a single model owns both extremes}:
Qwen3-8B scores $0.314$ on synthetic and $0.947$ on Belebele. Within one model, across six
benchmarks, with decoder and task cap fixed, the floor swings $3\times$. This is the sharpest form
of the statement that no benchmark-level comparison on the raw metric is interpretable.

Writing $\kappa = (S-c)/(1-c)$, chance-corrected retention is $15$--$18\%$ rather than $71$--$73\%$
chance-inclusive: of the agreement a model achieves above chance when asked the same question
twice in one language, roughly one sixth survives a change of language. Every figure in this
appendix is computed on the \emph{macro average of per-cell ratios}, which is why the frontier band
here is $3.0$ points rather than the $2.6$ of the pooled aggregation the main text uses; the three
aggregations are reconciled in Appendix~\ref{app:scale}. On that basis the band is $3.0$ absolute
points before correction and $3.0$ after. \textbf{In relative terms it widens, from
$4.1\%$ to $18.8\%$}, because the correction shrinks the level roughly fivefold while leaving the
between-model differences nearly intact; we report both metrics rather than the absolute one
alone. The measured floor is \emph{model-specific} (Qwen3-235B $0.450$ against Gemma-3-27B $0.623$, since Qwen3's
traces are longest), and those differences very nearly cancel the differences in raw $\tilde{I}$.
A sensitivity analysis that assumes one uniform $c$ for every model therefore misestimates the
risk: it predicts the band breaks near $c \approx 0.3$, whereas at the measured $c \approx 0.56$
the band is unchanged.

No cell has $I_{\text{cross}} \leq c$, so no ratio is sign-flipped or undefined; the tightest
margin in the set is Qwen3-8B on Belebele at $+0.0024$. Corrected values span $0.023$--$0.689$
with no divergence. The correction divides by $1-c$, so cells with a very high floor are
numerically delicate: at $c = 0.9473$ the divisor is $0.0527$, and a $0.005$ error in $c$ would
move that cell's corrected value from $+0.046$ to $-0.055$. \textbf{No conclusion rests on such a
cell}: Qwen3-8B falls outside the corrected band on its pooled value across all six of its cells,
not on Belebele alone.

\section{Scale and Vendor Generalisation}
\label{app:scale}

Three systems were chosen so that a break in the frontier regularity would be interpretable rather
than merely observed: \textbf{Gemma-3-4B} holds family and training recipe fixed against
Gemma-3-27B at $6.75\times$ scale difference; \textbf{Qwen3-8B} does the same for Qwen at
$29\times$; \textbf{Aya-Expanse-8B} is an independent vendor with explicitly multilingual
post-training, the rung most likely to break a training-recipe account. Each was run for two
replicates at $T{=}0$ on the 300-task subset (precisely the grid cell that produced the
$71$--$73\%$ figure), for $36$ cells and $135{,}600$ rollouts.

Pooled over pairs, $\tilde{I}$ is Qwen3-8B $0.8016$ $[0.7932, 0.8092]$, Gemma-3-4B $0.6259$
$[0.6192, 0.6323]$ and Aya-Expanse-8B $0.4738$ $[0.4658, 0.4811]$, against the frontier band
$[0.708, 0.733]$. \textbf{At least two of the three fall outside the band on every treatment we
tried}: three of three under pooled-raw, two of three under macro-of-ratios-raw, and two of three
chance-corrected. The minimum across treatments is two, which is the strongest form of the claim
that is true under all of them.

Chance correction is what makes the anti-scaling reading in \S\ref{sec:estimand} legible, so we
print those values rather than leave them to be inferred. The comparison below is on the
\emph{macro average of per-cell ratios} throughout, since that is the basis on which the floor is
measured (Appendix~\ref{app:chancefloor}); the raw figures therefore differ from the pooled ones
quoted elsewhere. Writing $\kappa = (S-c)/(1-c)$, the frontier band moves from $[0.713, 0.742]$ raw
to $[0.146, 0.176]$ corrected. Raw, Gemma-3-4B sits at $0.654$, \emph{below} the band, and Qwen3-8B
at $0.735$, \emph{inside} it. Corrected, that flips: Gemma-3-4B $\mathbf{0.155}$ lands
\emph{inside} the band and Qwen3-8B $\mathbf{0.109}$ falls \emph{below} it, with Aya-Expanse-8B at
$\mathbf{0.047}$. \textbf{The correction reverses which of the two smaller models looks better},
the opposite of what a scaling account predicts. The reversal also holds under the pooled
aggregation, where the raw values are $0.626$ and $0.802$.

Three aggregations of $\tilde{I}$ are defensible and we state which we use. Pooled over pairs
gives the band $[0.708, 0.733]$ (width $2.6$ pts); the ratio of macro-averaged cell means gives
$[0.713, 0.746]$ ($3.2$ pts); the macro average of per-cell ratios gives $[0.713, 0.742]$
($3.0$ pts). The \emph{width} is stable at $2.6$--$3.2$ points under all three, so the regularity
does not depend on the choice, but Gemma-3-27B and Qwen3-235B swap ends of the band between
pooled and macro, so no claim about which model retains most is made anywhere in the paper. The
main text uses the pooled aggregation throughout.

A cell in which most rollouts are empty measures adherence, not retention, so we re-pool over
cells that actually produced traces. For the two clean models the gate barely moves the estimate
and never toward the band: Gemma-3-4B $0.6259 \rightarrow 0.6307$ at $\leq5\%$ empty (5/6 cells
retained), Qwen3-8B $0.8016 \rightarrow 0.8094$ (5/6). For Aya the gate \emph{lowers} the estimate
to $0.4460$ and retains only 3 of 6 cells at $\leq20\%$ empty.

Aya-Expanse-8B is an adherence failure rather than a retention measurement. Its task-weighted empty-trace rate is $31.2\%$: FLORES-200 $86.3\%$, synthetic $82.1\%$, XNLI
$20.4\%$, XQuAD $14.0\%$, Belebele $11.9\%$, XCOPA $9.5\%$. On its two worst benchmarks the model
emits no parseable trace in more than four rollouts in five. We treat it exactly as GPT-OSS-120B
is treated in Appendix~\ref{app:gptoss}: reported, excluded from the regularity, and used as
evidence for the adherence story rather than the retention story. The independent-vendor question
therefore remains open.

Length does not explain the break. Mean trace length at $T{=}0$ is Qwen3-8B $1.99$, Gemma-3-27B $3.48$, Sarvam-M $3.88$, Llama-4
$4.07$, Gemma-3-4B $4.44$, Qwen3-235B $5.70$. Regressing $\tilde{I}$ on mean length across the six
protocol-adherent models gives $\tilde{I} = 0.817 - 0.0254\,\ell$ with $r = -0.547$,
$R^2 = 0.30$; per cell across 34 clean cells, $R^2 = 0.15$. The structural refutation is
Gemma-3-4B: at $4.44$ its mean trace length sits \emph{inside} the frontier range
$[3.48, 5.70]$, between Llama-4 and Qwen3-235B, and it still misses the band by $8.2$ points. A
model at frontier trace length with non-frontier retention is an observation the length hypothesis
cannot absorb.

Variance collapses with scale. Across the four frontier models at 17--235B the spread is $2.6$ points (SD $0.011$); across the
two clean models at 4--8B it is $17.6$ points (SD $0.088$): a $6.8\times$ difference in spread and
an $8.0\times$ difference in SD. We offer this as a hypothesis at $n = 2$ below 10B. Note also that both clean new models fall inside the
per-cell range $[0.61, 0.82]$ already reported for the frontier four: they break the
\emph{per-model} band while sitting within the \emph{per-cell} variation the study documents.
This is a raw-scale observation and is not comparable to the chance-corrected values above.

\section{Limitations}
\label{app:limitations}

What follows is what the study leaves open. Four questions that would otherwise belong here are
settled by experiment instead, and appear as results: trace-length causality
(Appendix~\ref{app:lengthcausal}), the chance floor (Appendix~\ref{app:chancefloor}), the head-room
account (Appendix~\ref{app:pivot}) and ceiling-corrected voting (Appendix~\ref{app:voting}).

Calls are parsed and compared but never dispatched, and no
observation is fed back into the loop. We therefore measure the \emph{induced} tool-use policy
under a controlled scaffold, not grounded execution. Whether the same divergence appears when
tool outputs change the state is untested and is the most important extension.

The frontier band rests on four systems, and the regime below it on two.
$\tilde{I} \in [0.708, 0.733]$ across four frontier models is a striking regularity, but $n = 4$,
and the reported intervals are task-level \emph{within} each model, so they establish that each
$\tilde{I}$ is precisely estimated rather than that four systems suffice to establish a constant.
The per-cell decomposition ($n = 24$, model identity $\eta^2 = 5.7\%$) is the stronger form of the
evidence and is what \S\ref{sec:estimand} relies on. Extending to eight models locates a boundary
but does not close the question: the claim that retention is variable below roughly 10B rests on
\textbf{two} clean models there (Gemma-3-4B and Qwen3-8B), which is a hypothesis, not a result. A
frontier model outside the band, or a sub-10B model inside it under chance correction, would bound
the regime differently. The uniformity is also specific to greedy decoding: at $T{=}0.5$ and
$T{=}0.7$ the ratio inflates toward 1 and the models spread apart, so the figure must always be
quoted with its temperature.

The independent-vendor question is open, because that vendor's model did not adhere.
Aya-Expanse-8B was chosen precisely as the rung most likely to break a training-recipe account of
the regularity, and it emits no parseable trace on $31.2\%$ of its rollouts, more than four in
five on its two worst benchmarks. We therefore report it as an adherence result and exclude it
from the retention comparison, exactly as GPT-OSS-120B is treated. The consequence is that all six
retention datapoints come from four vendors, and \emph{whether an explicitly multilingual
post-training recipe changes retention remains untested}. This is the single most consequential
gap in the boundary analysis, and it is a gap we created by our own eligibility rule rather than
one the data settled.

Trace length is causal but its sign is not universal. The three-level manipulation
establishes that $\tilde{I}$ moves $6$--$7$ points under a $3\times$ change in length, with
disjoint intervals, further than the entire across-model band. But the two models tested
\emph{disagree on the direction}: Qwen3 is monotone decreasing, Gemma is not, and its anomaly is
the shortest arm. A third model is needed before anything general can be said about which way
length pushes retention, and a four- or five-level ladder before the non-monotonicity can be
distinguished from noise. The instruction is also a soft prompt constraint rather than a decoding
limit, so the arms overlap in the tails.

The think-in-task-language arm was refused by both models
($<1\%$ compliance), so the designed contrast never existed. This must be reported as a
\emph{failed manipulation}, not as evidence that reasoning language does not matter. That
hypothesis remains untested, and the refusal itself is the finding.

The pivot ablation settles less than four models might suggest. The \emph{removal}
direction does not generalise. It lowers length-matched agreement on 4/6 benchmarks in three
models and on 1/6 in the fourth, which we read as dose--response, with the pivot load-bearing in
proportion to use, but that reading is inferred from four points rather than tested. The \emph{mandate}
direction was pre-registered and its \emph{ordering} confirmed, yet its \emph{magnitude} is
monotone in neither head-room nor movement, so we offer no prescription. Three specific weaknesses
travel with it. Sarvam-M's mandate reached only $81.4\%$ first-action compliance against
$98.7$--$99.6\%$ elsewhere, so part of its small magnitude may be weak manipulation. Head-room is
measured on \emph{first-action} \texttt{Translate} alone and does not capture pivoting later in a
trace or inside reasoning text, and the reasoning-language result above shows that internal
pivoting is prompt-resistant. And each arm has one seed and one replicate, so pooled differences
below roughly $\pm0.01$, which includes Sarvam's removal result, should be read as ``no effect''
rather than as a small positive one; no $\tilde{I}$ or gap is formable for these arms, only
$I_{\text{cross}}$.

Voting is measured at one temperature only, because it is undefined at $T{=}0$, where all
replicates coincide, so the ten-replicate experiment runs at $T{=}0.7$. The finding that voting
costs $1.6$--$1.9$ points of $\tilde{I}$ therefore \emph{cannot be checked at the temperature where
the retention regularity is defined}, and the intervals are disjoint partly because task-level
bootstraps within a model are narrow by construction. The effect is small in absolute terms, and a
practitioner whose objective is raw cross-language agreement rather than retention relative to a
model's own ceiling does obtain it ($+0.05$).

Finally, $S$ is a single sequence-similarity metric. The chance floor it lacks is now measured
rather than assumed (Appendix~\ref{app:chancefloor}), but the choice of $S$ itself is not varied:
whether an alternative trace-similarity family, such as edit distance under a tool-cost matrix or a
set-based measure insensitive to order, would preserve the frontier band is untested. Absolute
values of $I_{\text{cross}}$ should not be compared across metric definitions.

Gold exists for four of six benchmarks, and the
two excluded (FLORES-200, synthetic) are the longest-trace ones, so the correctness picture comes
from the shorter half of the suite. Scoring is deliberately lenient, which makes absolute
accuracies upper bounds; because the same scorer applies everywhere, cross-language comparisons
are unaffected. Invariance and accuracy are measured on the same rollouts but are not
interchangeable: among adherent models the association is $r \approx +0.38$ and it reverses in a
quarter of cells.

Sarvam-M has no $T{=}1.0$ rung on the temperature ladder, so the
flatness claim of \S\ref{sec:divergence} rests on the full five-point range in one model and four
points in the other; its think-natively arm covers one benchmark rather than six. Neither affects
a headline number, and both are visible in Appendices~\ref{app:ladder} and~\ref{app:pivot} rather
than absorbed into an average.

\section{Prompts}
\label{app:prompts}

The baseline scaffold is fixed across all languages and models:

\begin{quote}\ttfamily\scriptsize
You are an agent that solves tasks using tools.\\[2pt]
Available tools:\\
- Search(query: str) - Search for information\\
- Calc(expression: str) - Calculate mathematical expressions\\
- Translate(text: str) - Translate text to English\\
- Summarize(text: str) - Summarize text\\
- Finish(answer: str) - Provide final answer\\[2pt]
Rules:\\
- You must use tools for reasoning.\\
- Do NOT give direct answers without tools.\\
- Always follow this format:\\[2pt]
Thought: ...\\
Action: ToolName(arguments)\\[2pt]
Repeat until final answer using Finish().\\
Task: \{task\}
\end{quote}

Interventions modify exactly one element. \textbf{Tool removed}: \texttt{Translate} is deleted
from the tool list. \textbf{Tool mandated}: a rule is added requiring \texttt{Translate} as the
first action when the task is not in English. \textbf{Reason in English} / \textbf{reason
natively}: a rule is added requiring every \texttt{Thought:} line to be in English, or in the
task's language. \textbf{Few-shot}: two or four complete worked \texttt{Thought}/\texttt{Action}
examples are appended before the task. With no intervention set, the scaffold hashes
byte-identically to the baseline used in every run, which is verified programmatically.

\section{Compute}
\label{app:compute}

All generation ran on nodes of $8 \times$ NVIDIA B200 (192\,GB) under vLLM
0.18~\citep{kwon2023vllm}, with tensor/data-parallel layouts chosen per model
($1\times8$ for Gemma, Sarvam and GPT-OSS; $8\times1$ for Qwen3 and Llama-4). The main sweep
required 2h30m to 4h13m per model, and the full campaign is approximately 120 GPU-hours of B200
time. Every condition writes to an isolated, labelled results tree, and 53 automated checks assert
that no labelled run can be read as a baseline cell or resume from one. Analysis is CPU-only and
deterministic given a fixed bootstrap seed.

\section{Provenance}
\label{app:provenance}

An analysis that reads the wrong results directory produces a plausible number from incomplete
data, silently. Two properties therefore have to hold, and we check both rather than assume them:
every arm of the campaign must be exactly complete, and every reference to an arm must resolve to
exactly one directory. Table~\ref{tab:provenance} reports the outcome.

\begin{table}[!h]
\centering
\TableFontSize
\setlength{\tabcolsep}{6pt}
\renewcommand{\arraystretch}{1.08}
\begin{tabular}{@{}lr@{}}
\toprule
\hdr Check & Outcome \\
\midrule
\rw Arms present against planned & \textbf{44 / 44} across 21 conditions \\
Arms holding exactly 6 cells & \textbf{44 / 44} \\
\rw Arms holding exactly 22{,}600 rollouts & \textbf{44 / 44} \\
Total rollouts in the audited campaign & \textbf{994{,}400} \\
\rw Arms resolving to more than one directory & \goodcell{\textbf{0 of 44}} \\
Cells with $I_{\text{cross}} \leq c$ (chance floor) & \goodcell{\textbf{0 of 66}} \\
\bottomrule
\end{tabular}
\caption{Every arm is exactly complete and every reference resolves unambiguously. An ``arm'' is
one (condition, model) unit; 22{,}600 rollouts is $300$ tasks times the language count of each
benchmark, plus the two benchmarks with fewer than 300 available tasks run to completion.}
\label{tab:provenance}
\end{table}

Benchmarks are identified by content rather than by directory name. Run tags are length-capped, so
a long model identifier can push the benchmark suffix off the end of a cell directory name, and for
one model all five adapted-benchmark cells were named identically apart from their timestamp. A
resolver keyed on the name silently drops or collides them, so we key instead on each benchmark's
\emph{language count}, which is unique across the suite (FLORES-200 21, XQuAD 12, XNLI 15,
Belebele 16, XCOPA 11, synthetic 23) and survives any renaming. Under this resolver all main-sweep
results reproduce exactly ($\Delta = 0.0000$).

Incomplete generations are excluded structurally rather than by filtering: they are moved out of
the analysis tree before any script runs, and the scripts read only the active tree. The resolver
additionally raises on a repeated benchmark within a directory, so a duplicate cannot be silently
averaged even if one were left in place.

\end{document}